\documentclass[10pt,twocolumn,letterpaper]{article}

\usepackage{cvpr}
\usepackage{times}
\usepackage{epsfig}
\usepackage{graphicx}
\usepackage{amsmath}
\usepackage{amssymb}
\usepackage{multirow}
\usepackage{color}
\usepackage{booktabs}

\usepackage[pagebackref=true,breaklinks=true,letterpaper=true,colorlinks,citecolor=black,linkcolor=red,bookmarks=false]{hyperref}

\usepackage{algorithm}
\usepackage{algorithmic}
\cvprfinalcopy 

\def\cvprPaperID{1189} 

\ifcvprfinal\fi
\begin{document}

\title{Cascaded Deep Video Deblurring Using Temporal Sharpness Prior}

\author{Jinshan Pan, Haoran Bai, and Jinhui Tang\\
Nanjing University of Science and Technology\\
\\
}

\maketitle

\begin{figure}[t]\footnotesize
\vspace{-2.5in}
\begin{minipage}{\textwidth}
\centering
\begin{tabular}{cccccc}
\includegraphics[width=0.16\linewidth]{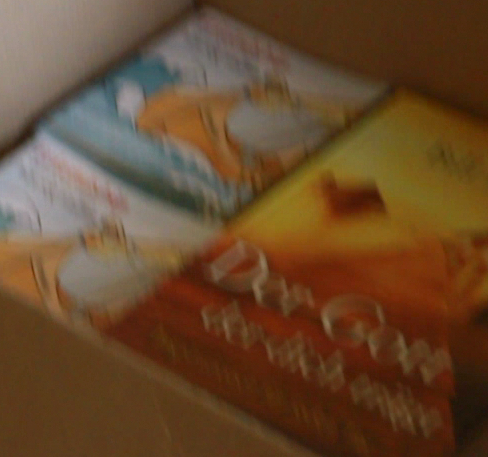}&\hspace{-4.5mm}
\includegraphics[width=0.16\linewidth]{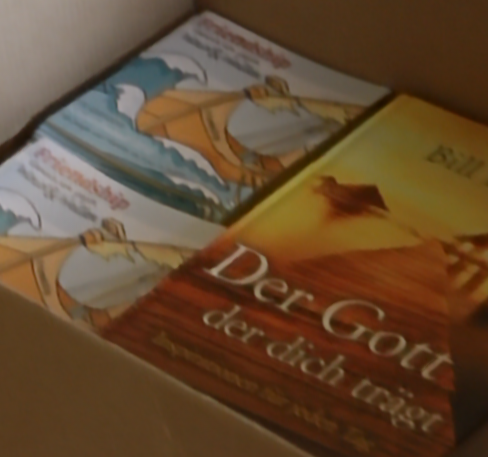} &\hspace{-4.5mm}
\includegraphics[width=0.16\linewidth]{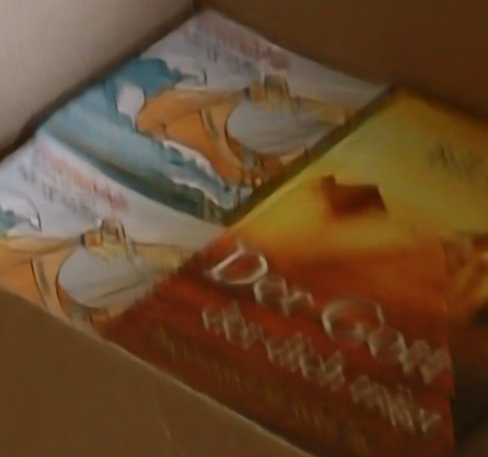} &\hspace{-4.5mm}
\includegraphics[width=0.16\linewidth]{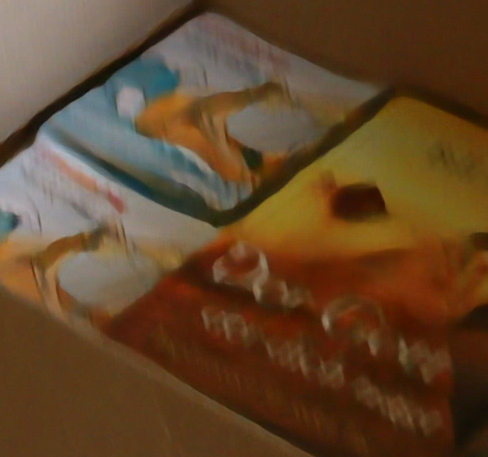} &\hspace{-4.5mm}
\includegraphics[width=0.16\linewidth]{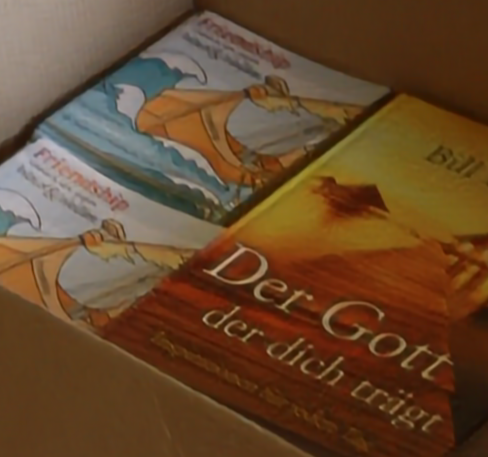} &\hspace{-4.5mm}
\includegraphics[width=0.16\linewidth]{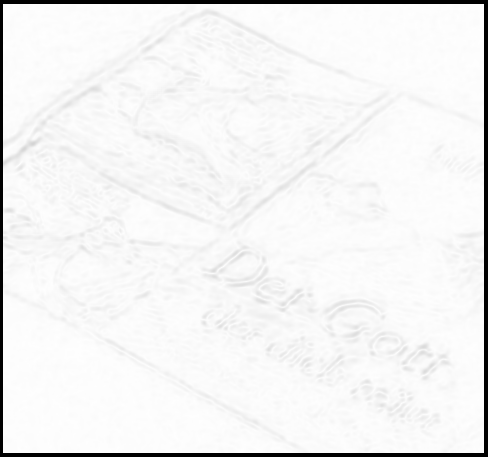} \\
(a) Input frame & \hspace{-0.5cm} (b) Kim and Lee~\cite{VD/kim/cvpr15} & \hspace{-0.5cm} (c) STFAN~\cite{zhoushanchen/iccv19} & \hspace{-0.5cm} (d) EDVR~\cite{edvr} & \hspace{-0.5cm} (e) Ours & \hspace{-0.5cm} (f) Sharpness prior of (a)\\
\end{tabular}
\caption{Deblurred result on a real challenging video. Our algorithm is motivated by the success of variational model-based methods.
It explores sharpness pixels from adjacent frames by a temporal sharpness prior (see (f)) and restores sharp videos by a cascaded inference process.
As our analysis shows, enforcing the temporal sharpness prior in a deep convolutional neural network (CNN) and learning the deep CNN by a cascaded inference manner can make the deep CNN more compact
and thus generate better-deblurred results than both the CNN-based methods~\cite{edvr,zhoushanchen/iccv19} and variational model-based method~\cite{VD/kim/cvpr15}.
%
}
\label{fig:teaser}
\end{minipage}
\end{figure}

\vspace{2mm}
\begin{abstract}
   We present a simple and effective deep convolutional neural network (CNN) model for video deblurring.
   %
   The proposed algorithm mainly consists of optical flow estimation from intermediate latent frames and latent frame restoration steps.
   It first develops a deep CNN model to estimate optical flow from intermediate latent frames and then restores the latent frames based on the estimated optical flow.
   To better explore the temporal information from videos, we develop a temporal sharpness prior to constrain the deep CNN model to help the latent frame restoration.
   We develop an effective cascaded training approach and jointly train the proposed CNN model in an end-to-end manner.
   We show that exploring the domain knowledge of video deblurring is able to make the deep CNN model more compact and efficient.
   Extensive experimental results show that the proposed algorithm performs favorably against state-of-the-art methods on the benchmark datasets as well as real-world videos.
   The training code and test model are available at \url{https://github.com/csbhr/CDVD-TSP}.
\end{abstract}

\section{Introduction}
\label{sec: introduction}

Video deblurring, as a fundamental problem in the vision and graphics communities, aims to estimate latent frames from a blurred sequence.
As more videos are taken using hand-held and onboard video capturing devices, this problem has received active research efforts within the last decade.
The blur in videos is usually caused by camera shake, object motion, and depth variation. Recovering latent frames is highly ill-posed as only the blurred videos are given.

To recover the latent frames from a blurred sequence, conventional methods usually make assumptions on motion blur and latent frames~\cite{VD/kim/cvpr15,variational/model/iccv07,cho/tog12/vd,kim/cvpr14/dynamic,shengyangdai/cvpr08,Wulff/eccv14/video/deblur}.
Among these methods, the motion blur is usually modeled as optical flow~\cite{VD/kim/cvpr15,variational/model/iccv07,shengyangdai/cvpr08,Wulff/eccv14/video/deblur}.
The key success of these methods is to jointly estimate the optical flow and latent frames under the constraints by some hand-crafted priors.
These algorithms are physically inspired and generate promising results.
However, the assumptions on motion blur and latent frames usually lead to complex energy functions which are difficult to solve.

The deep convolutional neural network (CNN), as one of the most promising approach, has been developed to solve video deblurring.
Motivated by the success of deep CNNs in single image deblurring, Su et al.~\cite{DVD/cvpr17} concatenate consecutive frames and develop a deep CNN based on an encoder-decoder architecture to directly estimate the latent frames.
Kim et al.~\cite{DTBN/kim/iccv17} develop a deep recurrent network to recurrently restore latent frames by the concatenating multi-frame features.
To better capture the temporal information, Zhang et al.~\cite{zhang/3d/convolution} develop spatial-temporal 3D convolutions to help latent frame restoration.
These methods perform well when the motion blur is not significant and displacement among input frames is small.
However, they are less effective for the frames containing significant blur and large displacement as they do not consider the alignment among input frames~\cite{details/devil}. 

To remedy this problem, several methods estimate the alignment among consecutive input frames explicitly~\cite{sstn/eccv18,Reblur2Deblur,edvr} or implicitly~\cite{zhoushanchen/iccv19}
to restore latent frames using end-to-end trainable deep CNNs.
For example, the alignment methods~\cite{STN} and~\cite{KPN/cvpr18} have been extended to handle video deblurring by~\cite{sstn/eccv18} and~\cite{zhoushanchen/iccv19}.
The methods by~\cite{Reblur2Deblur,edvr} explicitly adopt the optical flow or deformable convolution to estimate the alignment among consecutive input frames to help video deblurring.
These algorithms show that using better alignment strategies is able to improve the performance of video deblurring.
Nevertheless, the main success of these algorithms is due to the use of large-capacity models. These models cannot be generalized well on real cases.
We note there exist lots of prior knowledge in variational model-based approaches and have been effective in video deblurring.
A natural question is that can we use the domain knowledge in variational model-based approaches to make deep CNN models more compact so that they can improve the accuracy of video deblurring?

To solve this problem, we propose a simple and compact CNN model for video deblurring.
%
Different from the variational model-based methods that warp the consecutive frames to generate blurred frames based on the estimated optical flow,
our algorithm warps the adjacent frames into the reference frame so that the consecutive frames align well and thus generating a clearer intermediate latent frame.
As the generated intermediate latent frame may contain artifacts and blur effect, we further develop a deep CNN model based on an encoder-decoder architecture to remove artifacts and blur.
To better explore the properties of consecutive frames, we develop a temporal sharpness prior to constrain the deep CNN models.
However, as our algorithm estimates optical flow from intermediate latent frames as the motion blur information, it requires a feedback loop.
To effectively train the proposed algorithm, we develop a cascaded training approach and jointly train the proposed model in an end-to-end manner.
Extensive experiments show that the proposed algorithm is able to generate favorable results against state-of-the-art methods as shown in Figure~\ref{fig:teaser}.

The main contributions are summarized as follows:

\begin{itemize}
  \item We propose a simple and compact deep CNN model that simultaneously estimates the optical flow and latent frames for video deblurring.
  \item To better explore the properties of consecutive frames, we develop a temporal sharpness prior to constrain deep CNN models.
  \item We quantitatively and qualitatively evaluate the proposed algorithm on benchmark datasets and real-world videos and show that it performs favorably against state-of-the-art methods in terms of accuracy and model size.
\end{itemize}

\section{Related Work}
\label{sec: Related-Work}
\noindent{\bf Hand-crafted prior-based methods.}
Early video or multi-frame deblurring methods~\cite{cho/tog12/vd,lucky/frames/06} usually assume that there exist sharp contents and interpolate them to help the restoration of latent frames.
The main success of these methods is due to the use of sharp contents from adjacent frames.
However, these methods are less effective for the blur caused by moving objects and usually generate smooth results due to the interpolation.

To overcome this problem, several algorithms~\cite{shengyangdai/cvpr08,kim/cvpr14/dynamic,VD/kim/cvpr15,li/cvpr10/vd} formulate the video deblurring by a variational approach.
These algorithms first formulate motion blur as optical flow and develop kind of priors to constrain the latent frames and optical flow for video deblurring.
Dai and Wu~\cite{shengyangdai/cvpr08} analyze the relations of motion blur and optical flow and alternatively estimate the transparency map, foreground, and background of latent frames.
As this method relies on the accuracy of transparency maps, it is further extended by~\cite{kim/cvpr14/dynamic}, where deblurring process is achieved by alternatively estimating optical flow and latent frames.
Kim et al.~\cite{VD/kim/cvpr15} approximate the motion blur using bidirectional optical flows based on~\cite{kim/cvpr14/dynamic}.
%
To deal with more complex motion blur, Gong et al.~\cite{Gong/flow/blur} develop CNNs to estimate optical flow and use the conventional deconvolution algorithm~\cite{epll} to restore latent frames.
In~\cite{Wulff/eccv14/video/deblur}, Wulff and Black develop a novel layered model of scenes in motion and restore latent frames layer by layer.
These algorithms are based on the physics models, which are able to remove blur and generate decent results.
However, the priors imposed on motion blur and latent frames usually lead to complex energy functions which are difficult to solve.
%

\noindent{\bf Deep learning based-methods.}
Due to the success of CNNs based on encoder and decoder architectures in image restoration~\cite{Encoders/inpainting,encoders/restoration},
this kind of network has been widely used in multi-frame~\cite{burst/deblurring} or video deblurring~\cite{DVD/cvpr17}.
Instead of using 2D convolution, Zhang et al.~\cite{zhang/3d/convolution} employ spatial-temporal 3D convolutions to help latent frame restoration.
As demonstrated by~\cite{details/devil}, these methods can be improved using optical flow for alignment.
To better use spatial and temporal information, Kim et al.~\cite{sstn/eccv18} develop an optical flow estimation step for alignment and aggregate information across the neighboring frames to restore latent ones.
Wieschollek et al.~\cite{Wieschollek/vd/iccv17} recurrently use the features from the previous frame in multiple scales based on a recurrent network.
In~\cite{DTBN/kim/iccv17}, Kim et al. develop a spatial-temporal recurrent network with a dynamic temporal blending layer for latent frame restoration.
Zhou et al. extend the kernel prediction network~\cite{KPN/cvpr18} to improve frame alignment.
In~\cite{edvr}, Wang et al., develop pyramid, cascading, and deformable convolution to achieve better alignment performance.
The latent frames are restored by a deep CNN model with temporal and spatial attention strategies.
By training the networks in an end-to-end manner, these aforementioned methods generate promising deblurred results.

We note that the main success of these algorithms on video deblurring is due to the use of large-capacity models.
Their generalization ability on real applications is limited as shown in Figure~\ref{fig:teaser}.
Different from these methods, we explore the simple and well-established principles to make the CNN model more compact instead of enlarging network model capacity for video deblurring.
%

%

\section{Motivation}
\label{sec: motivation}
To better motivate our work, we first revisit the conventional variational model-based methods.

For the blur process in videos, the $i$-th blurred image is usually modeled as:
\begin{equation}
\mathbf{B}_i = \frac{1}{2\tau}\int_{t = 0}^{\tau}\mathcal{H}_{i\to i+1}^{t}(\mathbf{I}_i) + \mathcal{H}_{i\to i-1}^{t}(\mathbf{I}_i)dt,
\label{eq: video-blur-model}
\end{equation}
where $\mathbf{I}_i$ denotes the $i$-th clear image; $\tau$ denotes the relative exposure time (which also means the camera duty cycle);
$\mathcal{H}_{i\to i+1}^{t}$ and $\mathcal{H}_{i\to i-1}^{t}$ denote the warping functions which warp the frame $\mathbf{I}_i$ into $\mathbf{I}_{i+1}$ and $\mathbf{I}_{i-1}$.
If we denote the bidirectional optical flow at frame $i$ as $\mathbf{u}_{i\to i+1}$ and $\mathbf{u}_{i\to i-1}$,
$\mathcal{H}_{i\to i+1}^{t}(\mathbf{I}_i)$ and $\mathcal{H}_{i\to i-1}^{t}(\mathbf{I}_i)$ can be represented as $\mathbf{I}_i(\mathbf{x} + t\mathbf{u}_{i\to i+1})$ and $\mathbf{I}_i(\mathbf{x} + t\mathbf{u}_{i\to i-1})$.

Based on the blur model~\eqref{eq: video-blur-model}, the deblurring process can be achieved by minimizing:
\vspace{-1mm}
\begin{align}
&\mathcal{L}(\mathbf{u}, \mathbf{I}) = \sum_i\rho_I(\mathcal{W}(\mathbf{I}_i), \mathbf{B}_i) + \varphi(\mathbf{I}_i) \\\nonumber
& + \sum_i\sum_{j}\rho_u\left(\mathbf{I}_i, \mathbf{I}_{i+j}(\mathbf{x} + \mathbf{u}_{i\to i+j})\right) + \phi(\mathbf{u}_{i\to i+j}),
\label{eq: video-deblur-model}
\vspace{-1mm}
\end{align}
where $\rho_I(\mathcal{W}(\mathbf{I}_i), \mathbf{B}_i)$ denotes the data term w.r.t. $\mathcal{W}(\mathbf{I}_i)$ and $\mathbf{B}_i$; $\mathcal{W}(\mathbf{I}_i)$ denotes the integration term in~\eqref{eq: video-blur-model};
 $\rho_u\left(\mathbf{I}_i(\mathbf{x}), \mathbf{I}_{i+j}(\mathbf{x} + \mathbf{u}_{i\to i+j})\right)$ denotes the data term w.r.t. $\mathbf{I}_i(\mathbf{x})$ and $\mathbf{I}_{i+j}(\mathbf{x} + \mathbf{u}_{i\to i+j})$;
 $\varphi(\mathbf{I}_i)$ and $\phi(\mathbf{u}_{i\to i+j})$ denote the constraints on latent image $\mathbf{I}_i$ and optical flow $\mathbf{u}_{i\to i+j}$.

In the optimization process, most conventional methods (e.g.,~\cite{VD/kim/cvpr15}) estimate the latent image and optical flow by iteratively minimizing:
\begin{equation}
\sum_i\rho_I(\mathcal{W}(\mathbf{I}_i), \mathbf{B}_i) + \varphi(\mathbf{I}_i),
\label{eq: video-deblur-model-image}
\end{equation}
and
\begin{equation}
\sum_i\sum_{j}\rho_u\left(\mathbf{I}_i, \mathbf{I}_{i+j}(\mathbf{x} + \mathbf{u}_{i\to i+j})\right) + \phi(\mathbf{u}_{i\to i+j}).
\label{eq: video-deblur-model-flow}
\vspace{-2mm}
\end{equation}
We note that alternatively minimizing~\eqref{eq: video-deblur-model-image} and~\eqref{eq: video-deblur-model-flow} is able to remove blur.
However, the deblurring performance mainly depends on the choice of constraints w.r.t. latent image $\mathbf{I}_i$ and optical flow $\mathbf{u}_{i\to i+j}$, and it is not trivial to determine proper constraints.
In addition, the commonly used constraints usually lead to highly non-convex objective functions which are difficult to solve.

We further note that most deep CNN-based methods directly estimate the sharp videos from blurred input and generate promising results.
However, they estimate the warping functions from blurred inputs instead of latent frames and do not explore the domain knowledge of video deblurring, which are less effective for the videos with significant blur effect.

To overcome these problems, we develop an effective algorithm which makes full use of the well-established principles in the variational model-based methods and explores the domain knowledge
to make deep CNNs more compact for video deblurring.

\section{Proposed Algorithm}
The proposed algorithm contains the optical flow estimation module, latent image restoration module, and the temporal sharpness prior. The optical flow estimation module
provides motion information for the latent frame restoration, while the latent frame restoration module further facilitates the optical flow estimation so that it makes the estimated flow more accurate.
The temporal sharpness prior is able to explore the sharpness pixels from adjacent frames so that it can facilitate better frame restoration.
All the modules are jointly trained in a unified framework by an end-to-end manner.
%
%
In the following, we explain the main ideas for each component in details.
For simplicity, we use three adjacent frames to illustrate the main ideas of the proposed algorithm.
\subsection{Optical flow estimation}
\label{ssec: Optical flow estimation}
The optical flow estimation module is used to estimate optical flow between input adjacent frames, where the estimated optical flow provides the motion information for the image restoration~\eqref{eq: video-deblur-model-image}.
%
As demonstrated in~\cite{pwcnent/deqing}, the optical flow estimation~\eqref{eq: video-deblur-model-flow} can be efficiently solved by a deep neural network.
We use the PWC-Net~\cite{pwcnent/deqing} as the optical flow estimation algorithm.
%
Given any two intermediate latent frames $\mathbf{I}_{i}$ and $\mathbf{I}_{i+1}$, we compute optical flow by:
\vspace{-1mm}
\begin{equation}
\mathbf{u}_{i\to i+1} = \mathcal{N}_f(\mathbf{I}_{i}; \mathbf{I}_{i+1}),
\label{eq: optical-flow-estimation}
\end{equation}
where $\mathcal{N}_f$ denotes the optical flow estimation network which takes two images as the input. For any other two frames, the network $\mathcal{N}_f$ shares the same network parameters.%
%

\vspace{-1mm}
\subsection{Latent frame restoration}
\vspace{-1mm}
\label{sec: Latent frame restoration}
With the estimated optical flow, we can use variational model~\eqref{eq: video-deblur-model-image} to restore latent frames according to existing methods, e.g.,~\cite{VD/kim/cvpr15}.
However, solving~\eqref{eq: video-deblur-model-image} involves large computation of $\mathcal{W}(\mathbf{I}_i)$ and needs to define the prior on latent frame $\mathbf{I}_i$, which makes the restoration more complex.
We note that the effect of $\mathcal{W}(\mathbf{I}_i)$ (i.e., the blur process~\eqref{eq: video-blur-model}) is to generate a blurred frame so that it is closed to the observed input frame $\mathbf{B}_i$ as much as possible.
The discretization of~\eqref{eq: video-blur-model} can be written as~\cite{cho/tog12/vd}:
\vspace{-1mm}
\begin{equation}
\mathcal{W}(\mathbf{I}_i) = \frac{1}{1+2\tau}\sum_{d = 1}^{\tau}\left(\mathcal{H}_{i\to i+1}^{t}(\mathbf{I}_i)+\mathcal{H}_{i\to i-1}^{t}(\mathbf{I}_i) + \mathbf{I}_i(\mathbf{x})\right).
\label{eq: video-deblur-model-dis-0}
\end{equation}

According to the estimated optical flow $\mathbf{u}_{i\to i+1}$ and $\mathbf{u}_{i\to i-1}$, if we set $\tau$ to be $1$, $\mathcal{W}(\mathbf{I}_i)$ can be approximated by:
\vspace{-1mm}
\begin{equation}
\mathcal{W}(\mathbf{I}_i) = \frac{1}{3}\left(\mathbf{I}_i(\mathbf{x} + \mathbf{u}_{i\to i+1}) + \mathbf{I}_i(\mathbf{x} + \mathbf{u}_{i\to i-1}) + \mathbf{I}_i(\mathbf{x})\right).
\label{eq: video-deblur-model-dis-1}
\end{equation}
Instead of generating a blurred frame, we want to generate clear one according to the estimated optical flow $\mathbf{u}_{i-1\to i}$, and $\mathbf{u}_{i+1\to i}$ so that
$\mathbf{I}_{i+1}(\mathbf{x} + \mathbf{u}_{i+1\to i})$ and $\mathbf{I}_{i-1}(\mathbf{x} + \mathbf{u}_{i-1\to i})$ can be aligned with $\mathbf{I}_i(\mathbf{x})$ well.
Thus, we can use the following formula to update latent frame $\mathbf{{I}}_i$:
\vspace{-1mm}
\begin{equation}
\mathbf{{I}}_i \gets \frac{1}{3}\left(\mathbf{I}_{i+1}(\mathbf{x} + \mathbf{u}_{i+1\to i}) + \mathbf{I}_{i-1}(\mathbf{x} + \mathbf{u}_{i-1\to i}) + \mathbf{I}_i(\mathbf{x}) \right).
\label{eq: video-deblur-model-summarization}
\end{equation}
However, directly using~\eqref{eq: video-deblur-model-summarization} will lead to the results contains significant artifacts due to the misalignment from $\mathbf{I}_{i+1}(\mathbf{x} + \mathbf{u}_{i+1\to i})$ and $\mathbf{I}_{i-1}(\mathbf{x} + \mathbf{u}_{i-1\to i})$.
To avoid this problem and generate high-quality latent frame $\mathbf{I}_i$, we use $\mathbf{I}_{i+1}(\mathbf{x} + \mathbf{u}_{i+1\to i})$ and $\mathbf{I}_{i-1}(\mathbf{x} + \mathbf{u}_{i-1\to i})$
as the guidance frames and develop a deep CNN model to restore latent frame $\mathbf{{I}}_i$ by:
\begin{equation}
\mathbf{{I}}_i \gets \mathcal{N}_l(\mathcal{C}(\mathbf{I}_{i+1}(\mathbf{x} + \mathbf{u}_{i+1\to i}); \mathbf{I}_i(\mathbf{x});\mathbf{I}_{i-1}(\mathbf{x} + \mathbf{u}_{i-1\to i}))),
\label{eq: video-deblur-model-summarization-image-update}
\end{equation}
where $\mathcal{C}(\cdot)$ denotes the concatenation operation and $\mathcal{N}_l$ denotes the restoration network.
Similar to~\cite{pwcnent/deqing}, we use the bilinear interpolation to compute the warped frames.

For the deep CNN model $\mathcal{N}_l$, we use an encoder-decoder architecture based on~\cite{SRN}. However, we do not use the ConvLSTM module in $\mathcal{N}_l$.
Other network architectures are the same as~\cite{SRN}.
%
\subsection{Temporal sharpness prior}
\label{sec: Temporal sharpness prior}
As demonstrated in~\cite{cho/tog12/vd}, the blur in the video is irregular, and thus there exist some pixels that are not blurred. Following the conventional method~\cite{cho/tog12/vd},
we explore these sharpness pixels to help video deblurring.

According to the warped frames $\mathbf{I}_{i+1}(\mathbf{x} + \mathbf{u}_{i+1\to i})$ and $\mathbf{I}_{i-1}(\mathbf{x} + \mathbf{u}_{i-1\to i})$,
if the pixel $\mathbf{x}$ in $\mathbf{I}_{i}(\mathbf{x})$ is a sharp one, the pixel values of $\mathbf{I}_{i+1}(\mathbf{x} + \mathbf{u}_{i+1\to i})$ and $\mathbf{I}_{i-1}(\mathbf{x} + \mathbf{u}_{i-1\to i})$ should be close to that of $\mathbf{I}_{i}(\mathbf{x})$.
Thus, we define this criterion as:
\begin{equation}
\mathcal{S}_i(\mathbf{x}) = \exp\left(-\frac{1}{2}\sum_{j\&j\neq 0}\mathcal{D}(\mathbf{I}_{i+j}(\mathbf{x} + \mathbf{u}_{i+j\to i}); \mathbf{I}_{i}(\mathbf{x}))\right),
\label{eq: sharpeness-prior-1}
\end{equation}
where $\mathcal{D}(\mathbf{I}_{i+j}(\mathbf{x} + \mathbf{u}_{i+j\to i}); \mathbf{I}_{i}(\mathbf{x}))$ is defined as $\|\mathbf{I}_{i+j}(\mathbf{x} + \mathbf{u}_{i+j\to i})- \mathbf{I}_{i}(\mathbf{x})\|^2$.
Based on~\eqref{eq: sharpeness-prior-1}, if the value of $\mathcal{S}_i(\mathbf{x})$ is close to $1$, the pixel $\mathbf{x}$ is likely to be clear.
Thus, we can use $\mathcal{S}_i(\mathbf{x})$ to help the deep neural network to distinguish whether the pixel is clear or not so that it can help the latent frame restoration.
To increase the robustness of $\mathcal{S}_i(\mathbf{x}) $, we define $\mathcal{D}(\mathbf{I}_{i+j}(\mathbf{x} + \mathbf{u}_{i+j\to i}); \mathbf{I}_{i}(\mathbf{x}))$ as
\begin{align}
\mathcal{D}(\mathbf{I}_{i+j}(\mathbf{x} + &\mathbf{u}_{i+j\to i}); \mathbf{I}_{i}(\mathbf{x})) = \\\nonumber
&\sum_{\mathbf{y}\in \omega(\mathbf{x})}\left\|\mathbf{I}_{i+j}(\mathbf{y} + \mathbf{u}_{i+j\to i})- \mathbf{I}_{i}(\mathbf{y})\right\|^2,
\label{eq: sharpeness-prior-d}
\end{align}
where $\omega(\mathbf{x})$ denotes an image patch centered at pixel $\mathbf{x}$.
With the temporal sharpness prior $\mathcal{S}_i(\mathbf{x})$, we modify the latent frame restoration~\eqref{eq: video-deblur-model-summarization-image-update} by
\begin{equation}
\mathbf{{I}}_i \gets \mathcal{N}_l(\mathcal{C}(\mathcal{C}_{\mathbf{I}_{i}}; \mathcal{S}_i(\mathbf{x}))),
\label{eq: latent-image-sharpeness-prior-1}
\end{equation}
where $\mathcal{C}_{\mathbf{I}_{i}} = \mathcal{C}(\mathbf{I}_{i+1}(\mathbf{x} + \mathbf{u}_{i+1\to i}); \mathbf{I}_i(\mathbf{x});\mathbf{I}_{i-1}(\mathbf{x} + \mathbf{u}_{i-1\to i}))$.
We will show that using $\mathcal{S}_i(\mathbf{x})$ is able to help latent frame restoration.
\subsection{Inference}
As the proposed algorithm contains intermediate the optical flow estimation, latent frame estimation, and temporal sharpness computation, we train the proposed algorithm in a cascaded manner.

Let $\mathbf{\Theta}_t = \{\mathcal{O}_t, \mathcal{L}_t\}$ denote the model parameters of optical flow estimation and latent frame restoration networks at stage (iteration) $t$.
We learn the stage-dependent model parameters $\mathbf{\Theta}_t$ from $N$ training video sequences, where each video sequence contains~$\{\mathbf{B}_i^n, \mathbf{I}_{gt, i}^n\}_{i = 1}^{M}$ training samples.
Given $2j+1$ blurred frames, the parameter $\mathbf{\Theta}_t$ is learned by minimizing the cost function:
\begin{equation}
\label{eq: loss-function}
\mathcal{J}(\mathbf{\Theta}_t) = \sum_{n = 1}^{N}\sum_{i = 1}^{M} \|\mathcal{F}_{\mathbf{\Theta}_t}(\mathbf{B}_{i-j}^{n}; ...; \mathbf{B}_{i}^{n};...;\mathbf{B}_{i+j}^{n})-\mathbf{I}_{gt, i}^{n}\|_1,
\end{equation}
where $\mathcal{F}_{\mathbf{\Theta}_t}$ denotes the whole network for video deblurring, which takes $2j+1$ blurred frames as the input.
That is, the intermediate latent frame at $t$-stage is $\mathbf{I}_{i}^{t}= \mathcal{F}_{\mathbf{\Theta}_t}(\mathbf{B}_{i-j}^{n}; ...; \mathbf{B}_{i}^{n}; ...;\mathbf{B}_{i+j}^{n})$.

Algorithm~\ref{alg: cascaded-training-algorithm} summarizes the main steps of the cascaded training approach, where $T$ denotes the number of stages.
%

\begin{algorithm}[t]
\caption{Proposed cascaded training algorithm.}
\label{alg: cascaded-training-algorithm}
\begin{algorithmic}
\STATE {\textbf{Input:} Training video sequences~$\{\mathbf{B}_i^n, \mathbf{I}_{gt, i}^n\}_{i = 1}^{M}$; $n=1,..., N$.}
\STATE {Initialize $\mathbf{I}_i^{n} \gets \mathbf{B}_i^{n}$.}
\FOR {$t = 1 \to T$}
\FOR {Any three frames $\mathbf{I}_{i-1}^{n}$, $\mathbf{I}_{i}^{n}$, and $\mathbf{I}_{i+1}^{n}$}
\STATE {Estimating optical flow $\mathbf{u}_{i-1\to i}$, $\mathbf{u}_{i+1\to i}$ according to~\eqref{eq: optical-flow-estimation}.}
\STATE {Computing $\mathcal{S}_i(\mathbf{x})$ according to~\eqref{eq: sharpeness-prior-1}.}
\STATE {Latent frame restoration according to~\eqref{eq: latent-image-sharpeness-prior-1}.}
\ENDFOR
\STATE {Estimating model parameters $\mathbf{\Theta}_t$ by minimizing~\eqref{eq: loss-function}.}
\STATE {Updating $\mathbf{I}_i^{n}$ according to~\eqref{eq: latent-image-sharpeness-prior-1} with the estimated parameter $\mathbf{\Theta}_t$.}
\ENDFOR
\STATE \textbf{Output:} Model parameters $\{\mathbf{\Theta}_t\}_{t = 1}^{T}$.
\end{algorithmic}
\end{algorithm}

\section{Experimental Results}
\vspace{-1mm}
In this section, we evaluate the proposed algorithm using publicly available benchmark datasets and compare it to state-of-the-art methods.

\vspace{-2mm}
\subsection{Parameter settings and training data}
\label{ssec: Parameter settings and training data}
\vspace{-1mm}
For fair comparisons with state-of-the-art methods, we use the video deblurring dataset by Su et al.~\cite{DVD/cvpr17} for training and evaluation, where 61 videos are used for training and the remaining 10 videos for the test.
We use the similar data augmentation method to~\cite{zhoushanchen/iccv19} to generate training data. The size of each image patch is $256\times 256$ pixels.
We initialize the latent frame restoration network according to~\cite{kaiming/initialization} and train it from scratch.
For the PWC-Net, we use the pre-trained model~\cite{pwcnent/deqing} to initialize it.
In the training process, we use the ADAM optimizer~\cite{adam} with parameters $\beta_1 = 0.9$, $\beta_2 = 0.999$, and $\epsilon = 10^{-8}$.
The minibatch size is set to be $8$. The learning rates for $\mathcal{N}_l$ and PWC-Net are initialized to be $10^{-4}$ and $10^{-6}$ and decrease to half after every 200 epochs.
We empirically set $T = 2$ as a trade-off between accuracy and speed.
At each stage, we use $3$ frames to generate one deblurred image. Thus, the proposed algorithm needs $5$ frames when $T = 2$.
To better make the network compact, the network at each stage shares the same model parameters.
Similar to~\cite{hardminingloss/inpainting}, we further use the hard example mining strategy to preserve sharp edges.
%
We implement our algorithm based on the PyTorch. More experimental results are included in the supplemental material.
The training code and test model are available at the authors' website.
\vspace{-1mm}
\subsection{Comparisons with the state of the art}
\vspace{-1mm}
To evaluate the performance of the proposed algorithm, we compare it against state-of-the-art algorithms
including the variational model-based method~\cite{VD/kim/cvpr15} and deep CNNs-based methods~\cite{DVD/cvpr17,Gong/flow/blur,edvr,DTBN/kim/iccv17,sstn/eccv18,zhoushanchen/iccv19,SRN}.
%
To evaluate the quality of each restored image on synthetic datasets, we use the PSNR and SSIM as the evaluation metrics.

Table~\ref{tab: result-dvd-dataset} shows the quantitative results on the benchmark dataset by Su et al.~\cite{DVD/cvpr17},
where the proposed algorithm performs favorably against the state-of-the-art methods in terms of PSNR and SSIM.
%
\begin{table*}[t]
\vspace{-1mm}
  \caption{Quantitative evaluations on the video deblurring dataset~\cite{DVD/cvpr17} in terms of PSNR and SSIM.
  All the comparison results are generated using the publicly available code. All the restored frames instead of randomly selected 30 frames from each test set~\cite{DVD/cvpr17} are used for evaluations.
  }
   \vspace{1mm}
   \label{tab: result-dvd-dataset}
\footnotesize
\resizebox{\textwidth}{!}{
 \centering
 \begin{tabular}{lcccccccccccc}
    \toprule
    Methods          &Kim and Lee~\cite{VD/kim/cvpr15}   &Gong et al.~\cite{Gong/flow/blur}  &Tao et al.~\cite{SRN}   &Su et al.~\cite{DVD/cvpr17}   &Kim et al.~\cite{DTBN/kim/iccv17}   & EDVR~\cite{edvr} &STFAN~\cite{zhoushanchen/iccv19}       &Ours\\
    \hline
     PSNRs     &26.94                                  &28.27                              &29.98                    &30.01                          &29.95                                & 28.51                            &31.15                &\bf{32.13}    \\
     SSIMs     &0.8158                                 &0.8463                             &0.8842                   &0.8877                          &0.8692                              &0.8637                            &0.9049               &\bf{0.9268}     \\

 \bottomrule
  \end{tabular}
}
\vspace{-3mm}
\end{table*}
\begin{figure*}[!t]\footnotesize
\centering
\begin{tabular}{ccccc}

\includegraphics[width=0.245\linewidth]{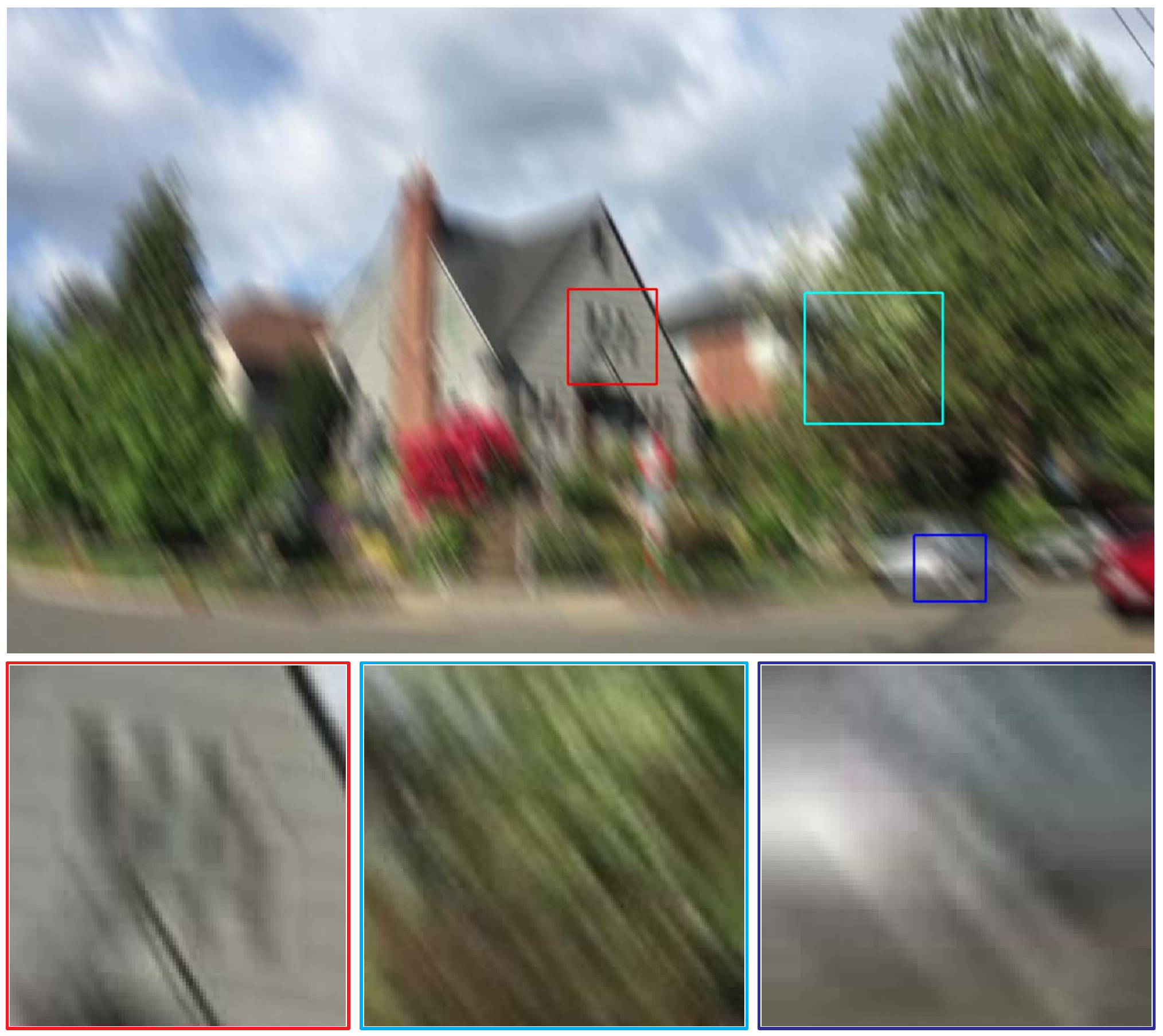} &\hspace{-4.5mm}
\includegraphics[width=0.245\linewidth]{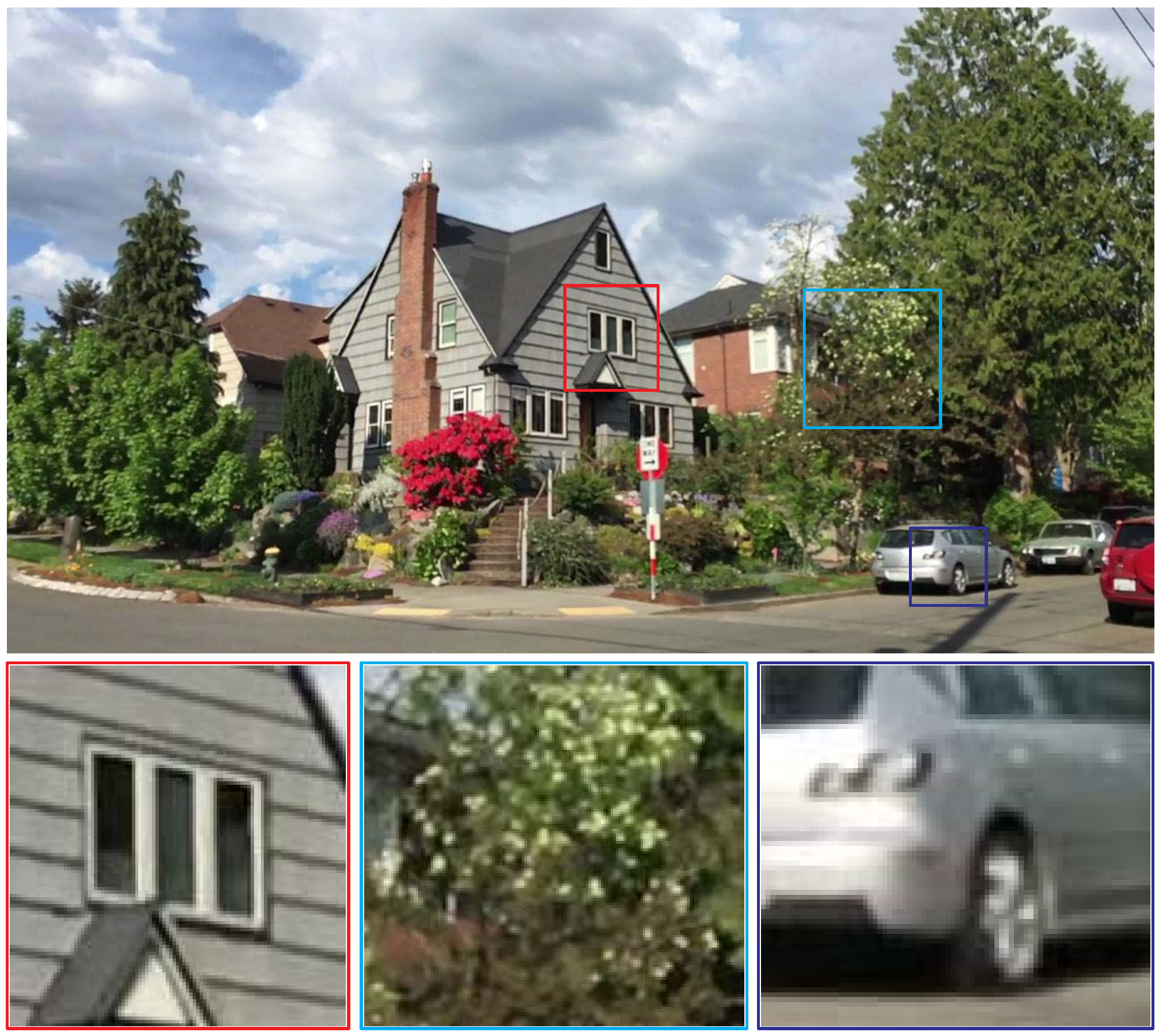} &\hspace{-4.5mm}
\includegraphics[width=0.245\linewidth]{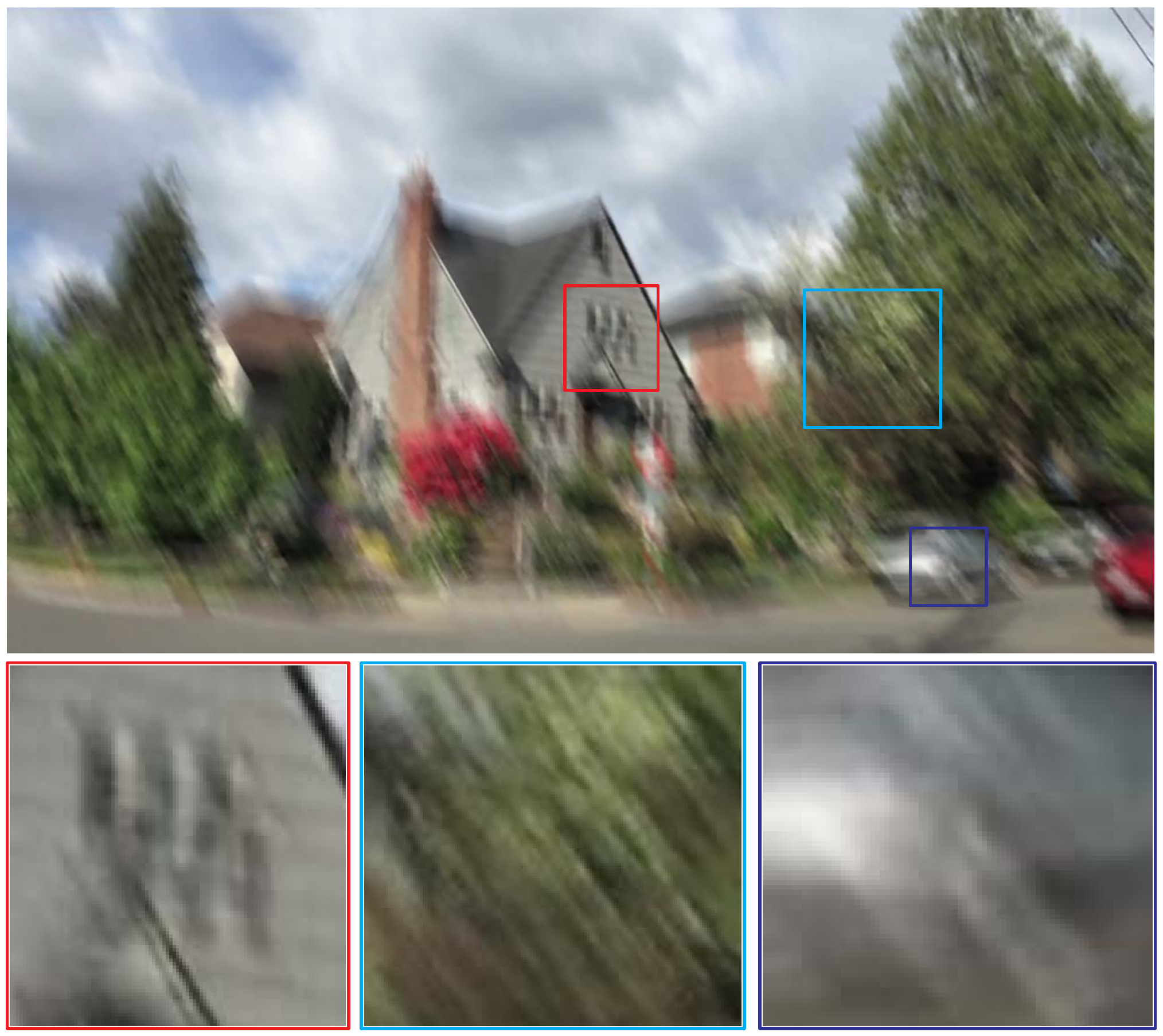} &\hspace{-4.5mm}
\includegraphics[width=0.245\linewidth]{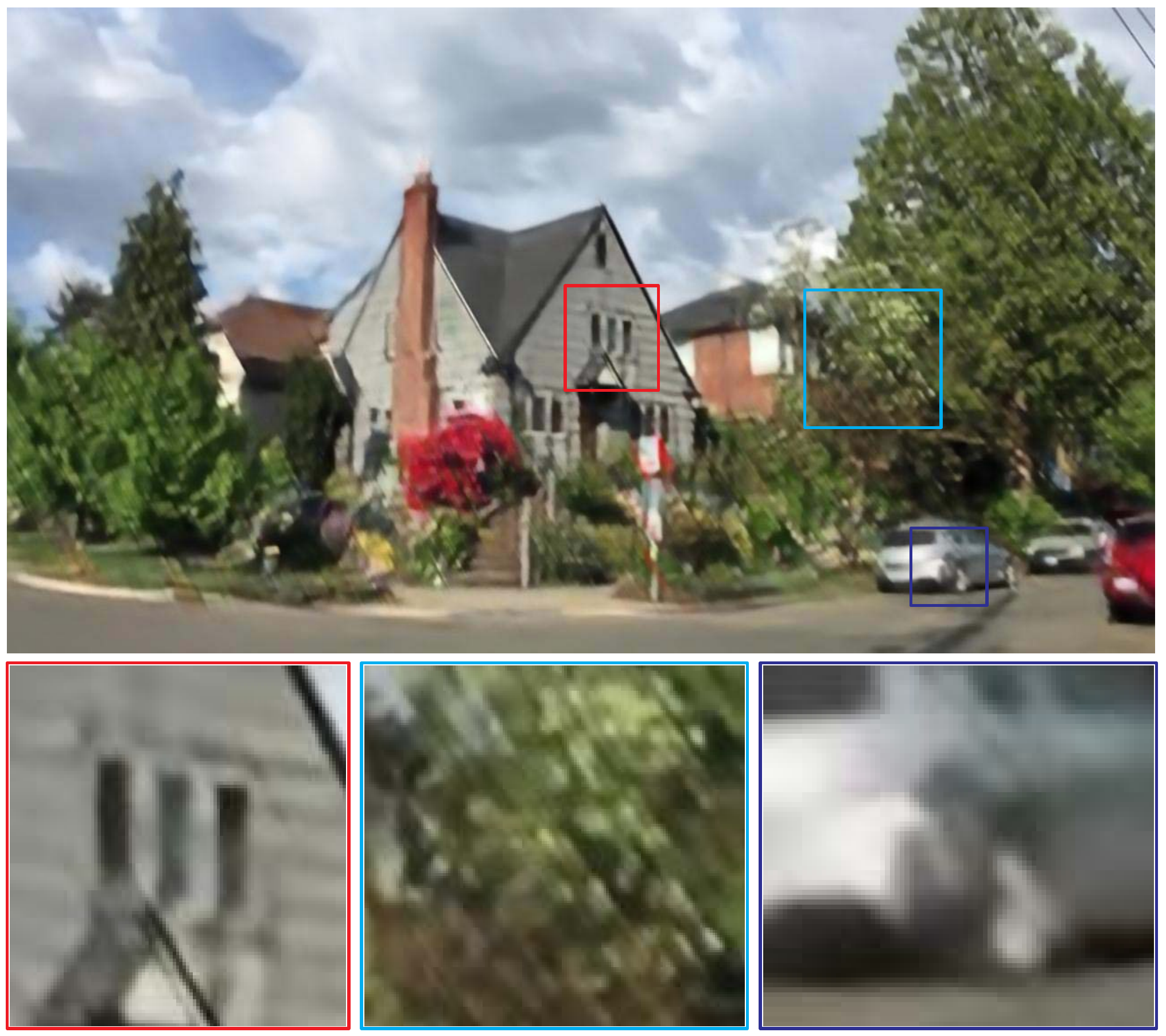} \\
(a) Blurred frame &\hspace{-4.5mm}  (b) GT &\hspace{-4.5mm}  (c) Kim and Lee~\cite{VD/kim/cvpr15}  &\hspace{-4.5mm}  (d) Tao et al.~\cite{SRN}\\
\includegraphics[width=0.245\linewidth]{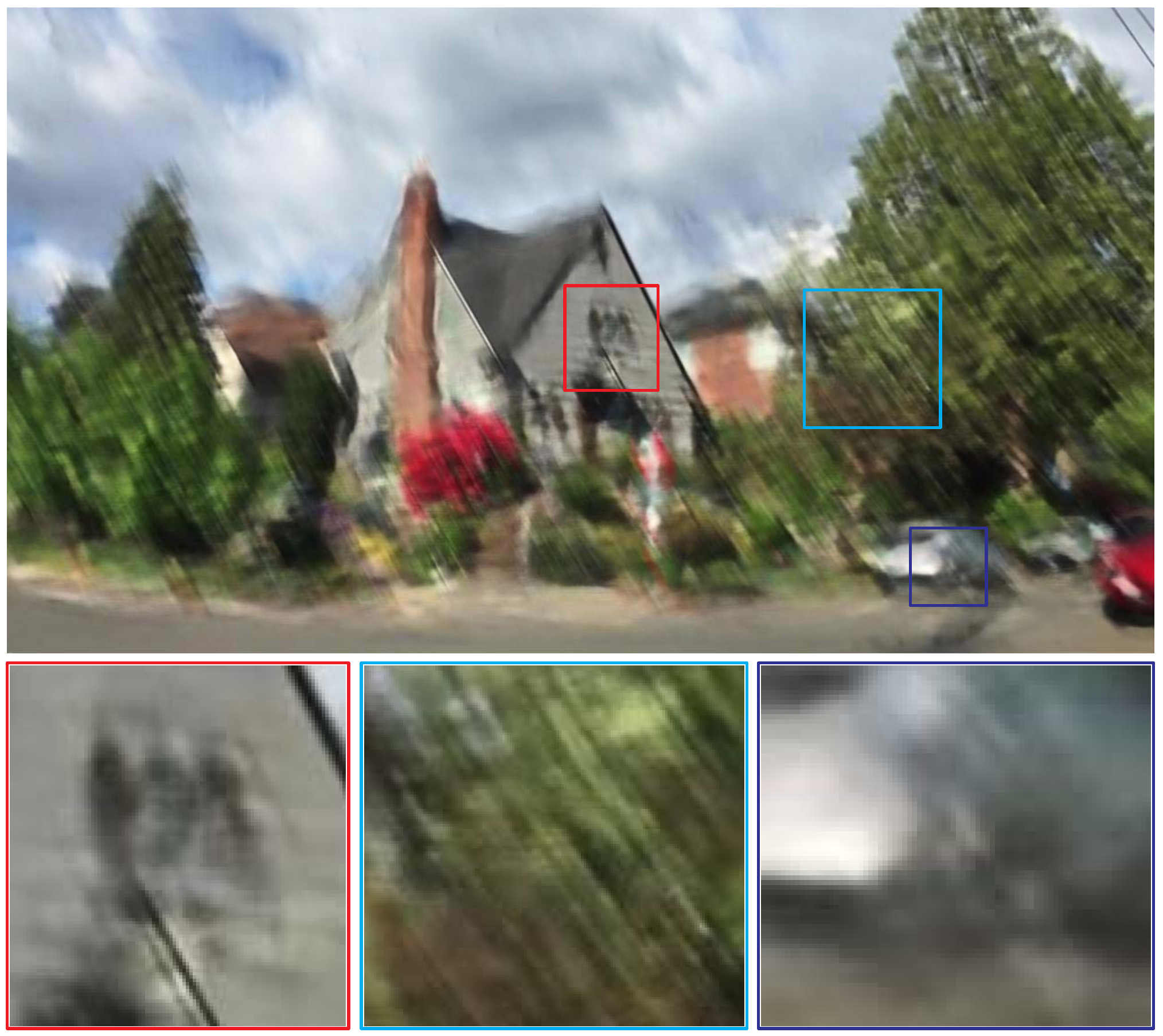}  &\hspace{-4.5mm}
\includegraphics[width=0.245\linewidth]{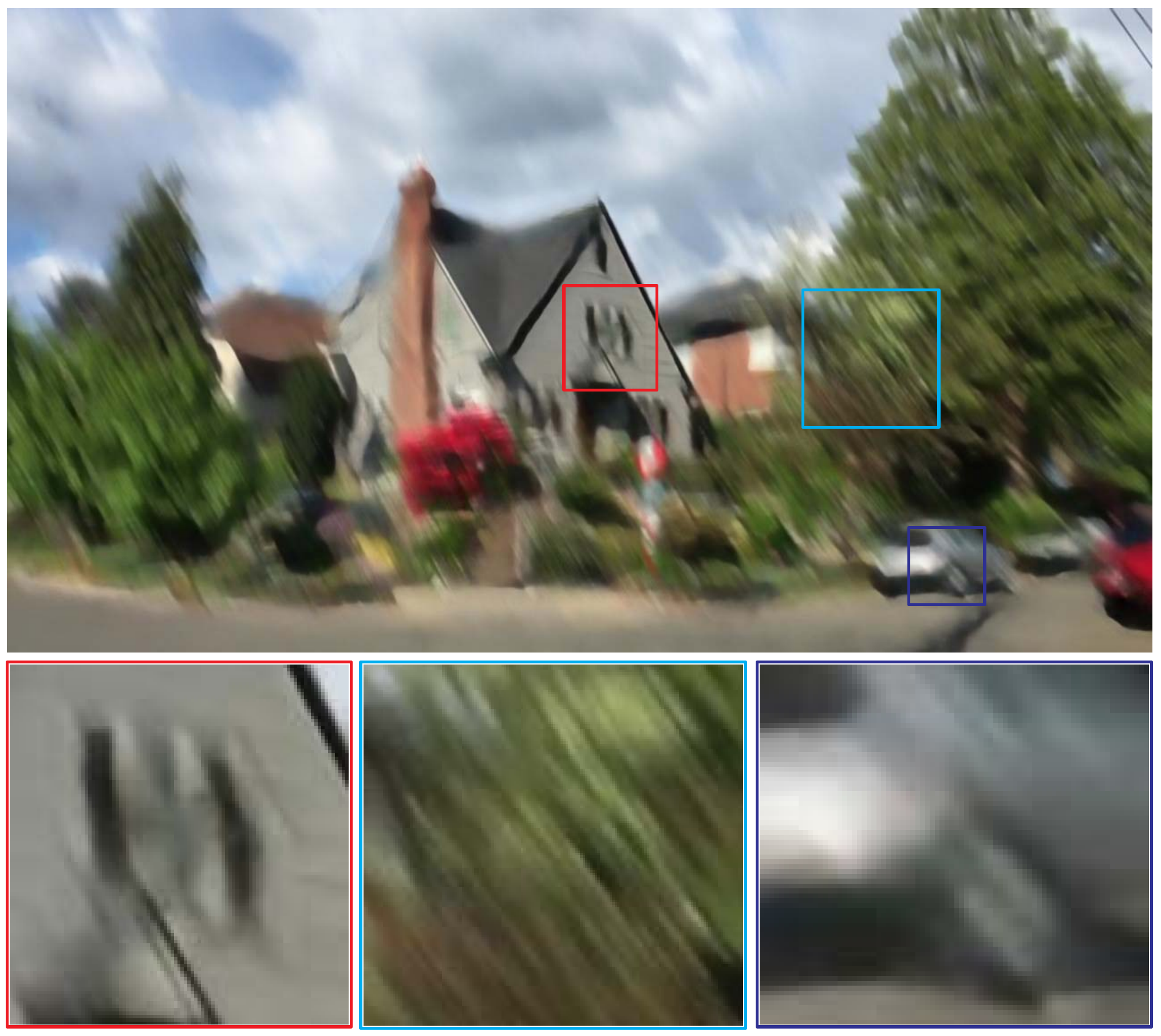} &\hspace{-4.5mm}
\includegraphics[width=0.245\linewidth]{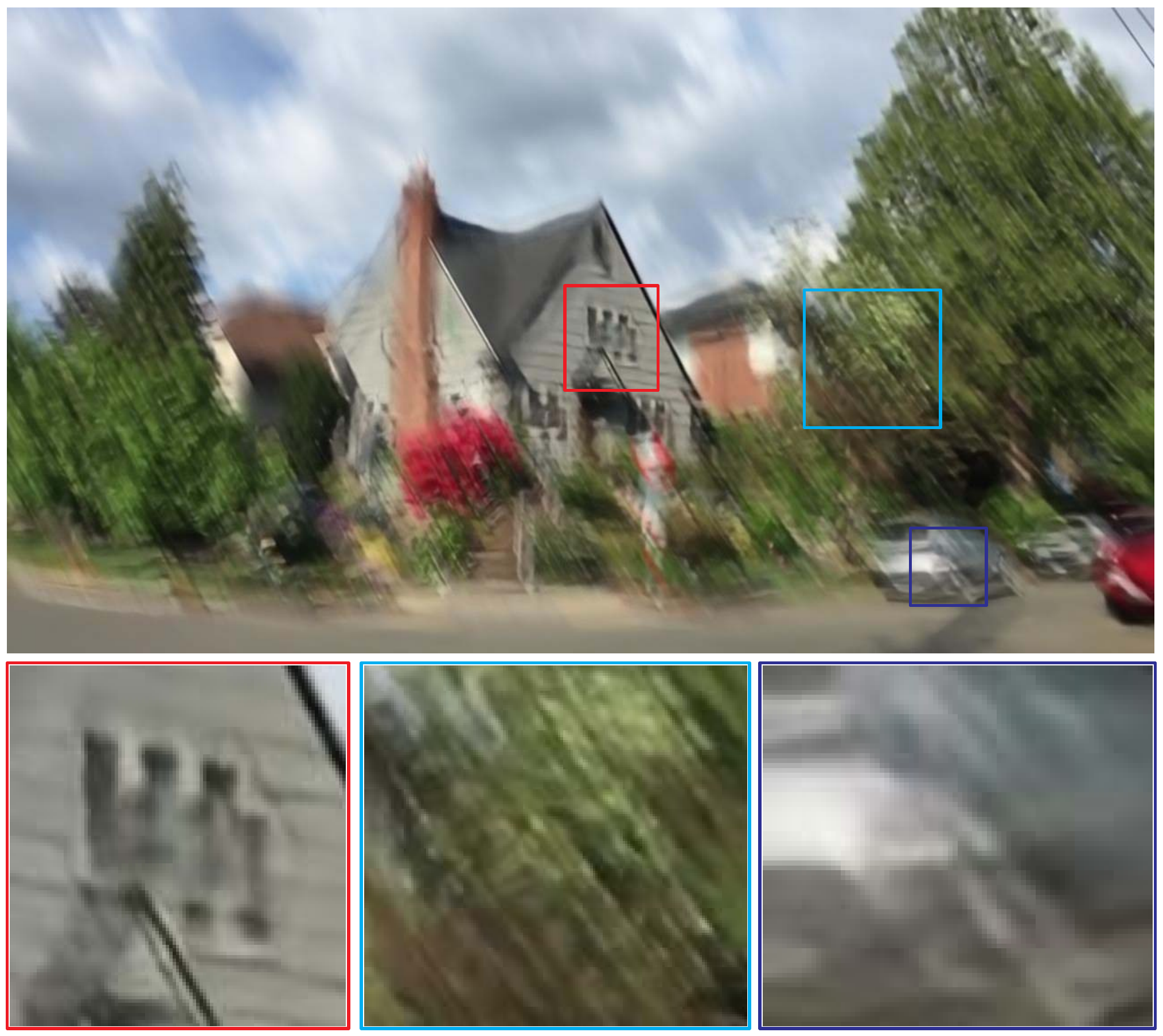} &\hspace{-4.5mm}
\includegraphics[width=0.245\linewidth]{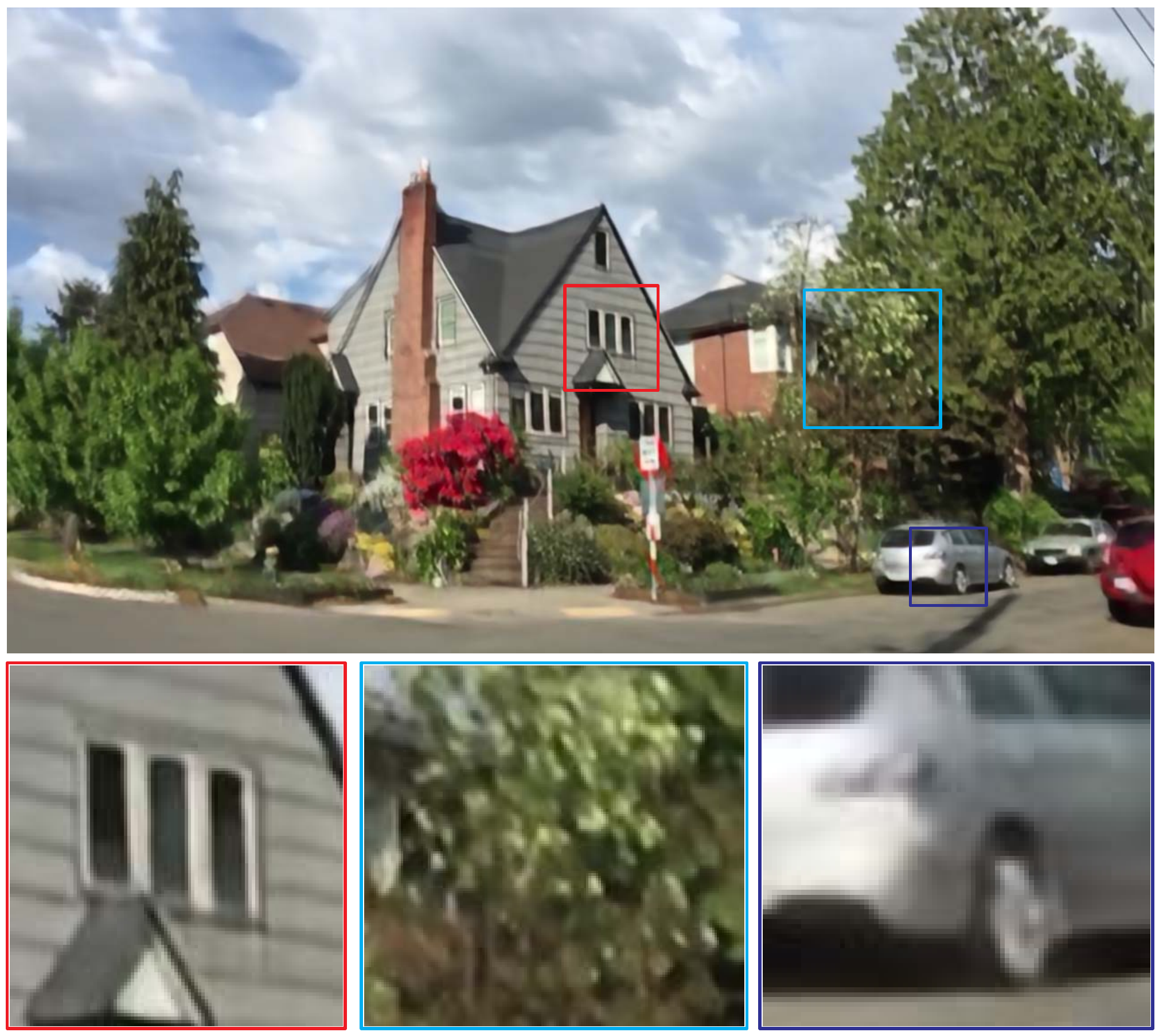}\\
(e) Su et al.~\cite{DVD/cvpr17} &\hspace{-4.5mm}  (f) EDVR~\cite{edvr} &\hspace{-4.5mm}  (g) STFAN~\cite{zhoushanchen/iccv19} &\hspace{-4.5mm}  (h) Ours\\
\end{tabular}
\caption{Deblurred results on the test dataset~\cite{DVD/cvpr17}. The deblurred results in (c)-(g) still contain significant blur effects. The proposed algorithm generates much clearer frames. }%
\label{fig: result-dvd-dataset-vis}
\vspace{-2mm}
\end{figure*}

Figure~\ref{fig: result-dvd-dataset-vis} shows some deblurred results from the test dataset~\cite{DVD/cvpr17}.
%
%
The variational model-based method~\cite{VD/kim/cvpr15} does not recover the structures well and generates the results with significant blur residual.
%
%
The method~\cite{SRN} develops end-to-end-trainable deep CNN models to deblur dynamic scenes. However, the deblurred images contain significant blur residual as the temporal information is not used.
%
%
The video deblurring algorithm~\cite{DVD/cvpr17} directly concatenates consecutive frames as the input of an end-to-end trainable deep CNN model.
However, the structures of the deblurred image are not sharp (Figure~\ref{fig: result-dvd-dataset-vis}(e)).
%
%
We note that the EDVR method~\cite{edvr} develops a pyramid, cascading, and deformable alignment module and uses a PreDeblur module for video deblurring.
However, this method is less effective when the PreDeblur does not remove blur from input frames.
The results in Figure~\ref{fig: result-dvd-dataset-vis}(f) show that the structures of the images by the EDVR method are not recovered well.
In contrast, the proposed method recovers finer image details and structures than the state-of-the-art algorithms.

We further evaluate the proposed method on the GOPRO dataset by Nah et al.~\cite{MSCNN/deblur/cvpr17}
following the protocols of state-of-the-art methods~\cite{DTBN/kim/iccv17,Nah/cvpr19}.
Table~\ref{tab: result-other-datasets} shows that the proposed algorithm generates the deblurred videos with higher PSNR and SSIM values.
\begin{table*}[t]
\vspace{-1mm}
  \caption{Quantitative evaluations on the video deblurring dataset~\cite{MSCNN/deblur/cvpr17} in terms of PSNR and SSIM. $*$ denotes the reported results from~\cite{Nah/cvpr19}.
  }
   \vspace{1mm}
   \label{tab: result-other-datasets}
\footnotesize
 \centering
 \begin{tabular}{lccccccccccccc}
    \toprule
    Methods                                             &Tao et al.~\cite{SRN}  &Su et al.~\cite{DVD/cvpr17}   &Wieschollek et al.~\cite{Wieschollek/vd/iccv17}$^*$         &Kim et al.~\cite{DTBN/kim/iccv17}$^*$    &Nah et al.~\cite{Nah/cvpr19}$^*$     &EDVR~\cite{edvr}      & STFAN~\cite{zhoushanchen/iccv19}         &Ours\\
    \hline
                                                        PSNRs        &30.29    &27.31                          & 25.19                &26.82          &29.97             &26.83       &28.59         &\bf{31.67}    \\
                                                        SSIMs        &0.9014   &0.8255                         &0.7794                &0.8245         &0.8947            &0.8426       &0.8608        &\bf{0.9279}     \\
 \bottomrule
  \end{tabular}
\vspace{-1mm}
\end{table*}
\begin{figure*}[!t]\footnotesize
\centering
\begin{tabular}{ccccc}
\includegraphics[width=0.245\linewidth]{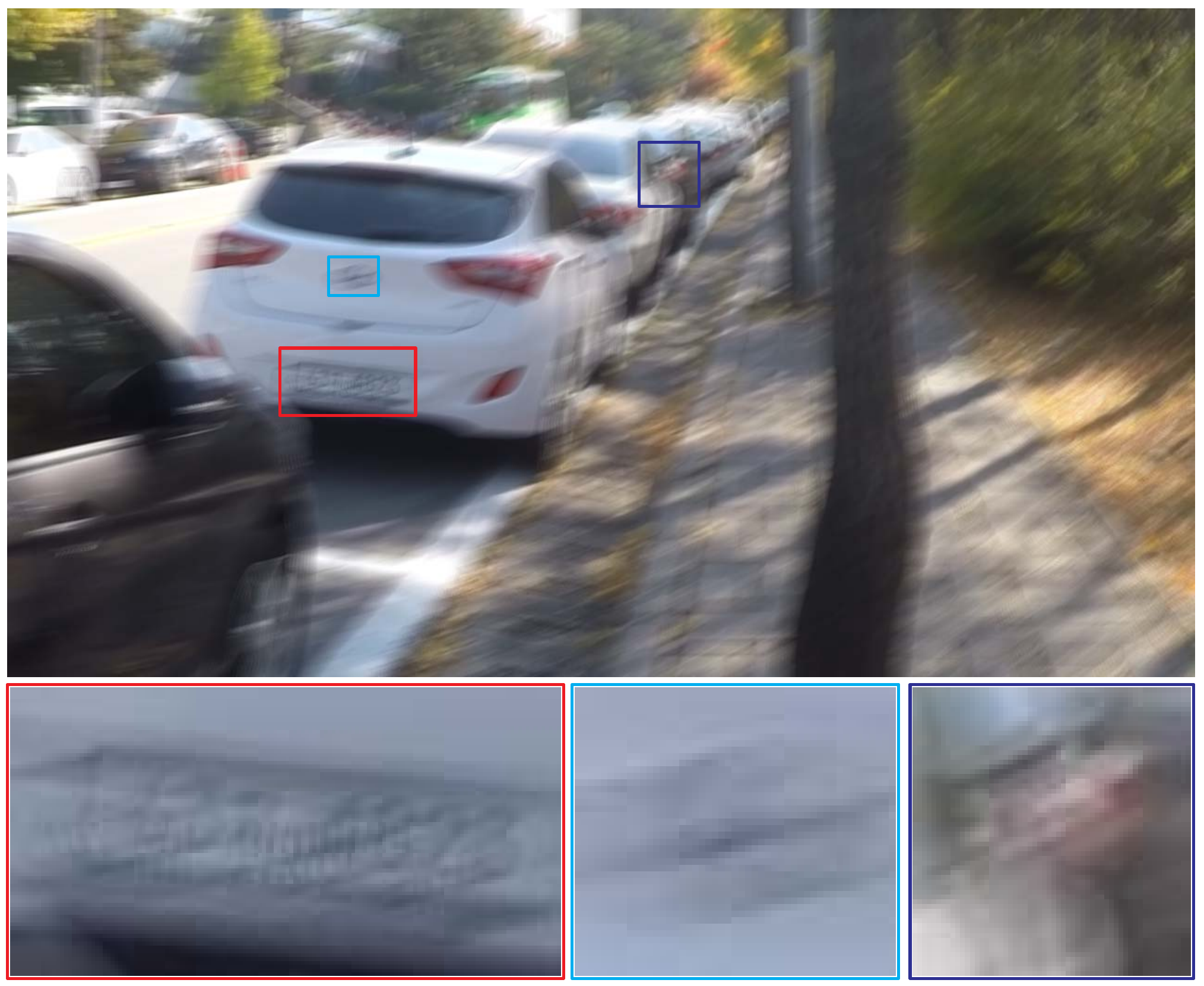} &\hspace{-4.5mm}
\includegraphics[width=0.245\linewidth]{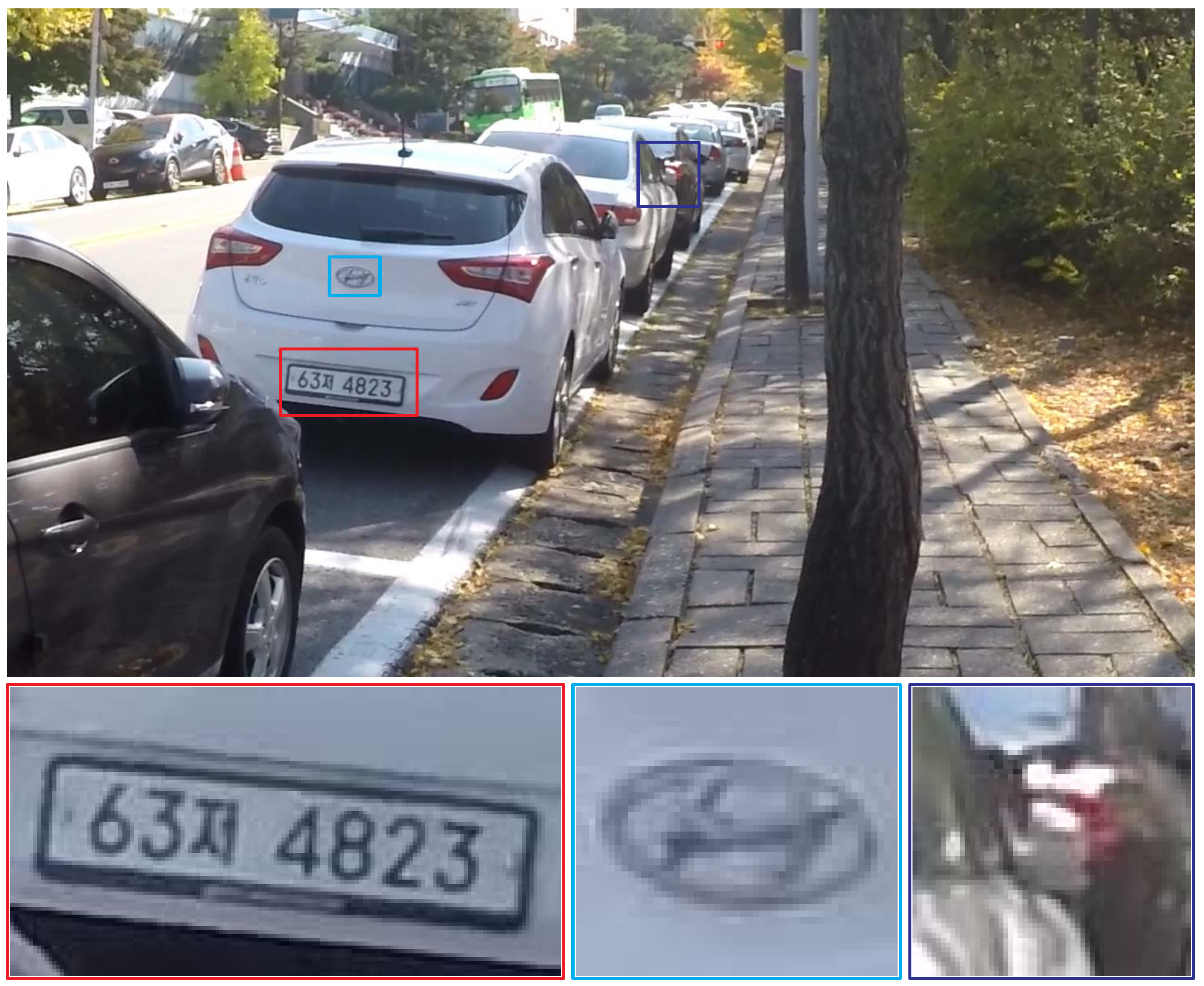} &\hspace{-4.5mm}
\includegraphics[width=0.245\linewidth]{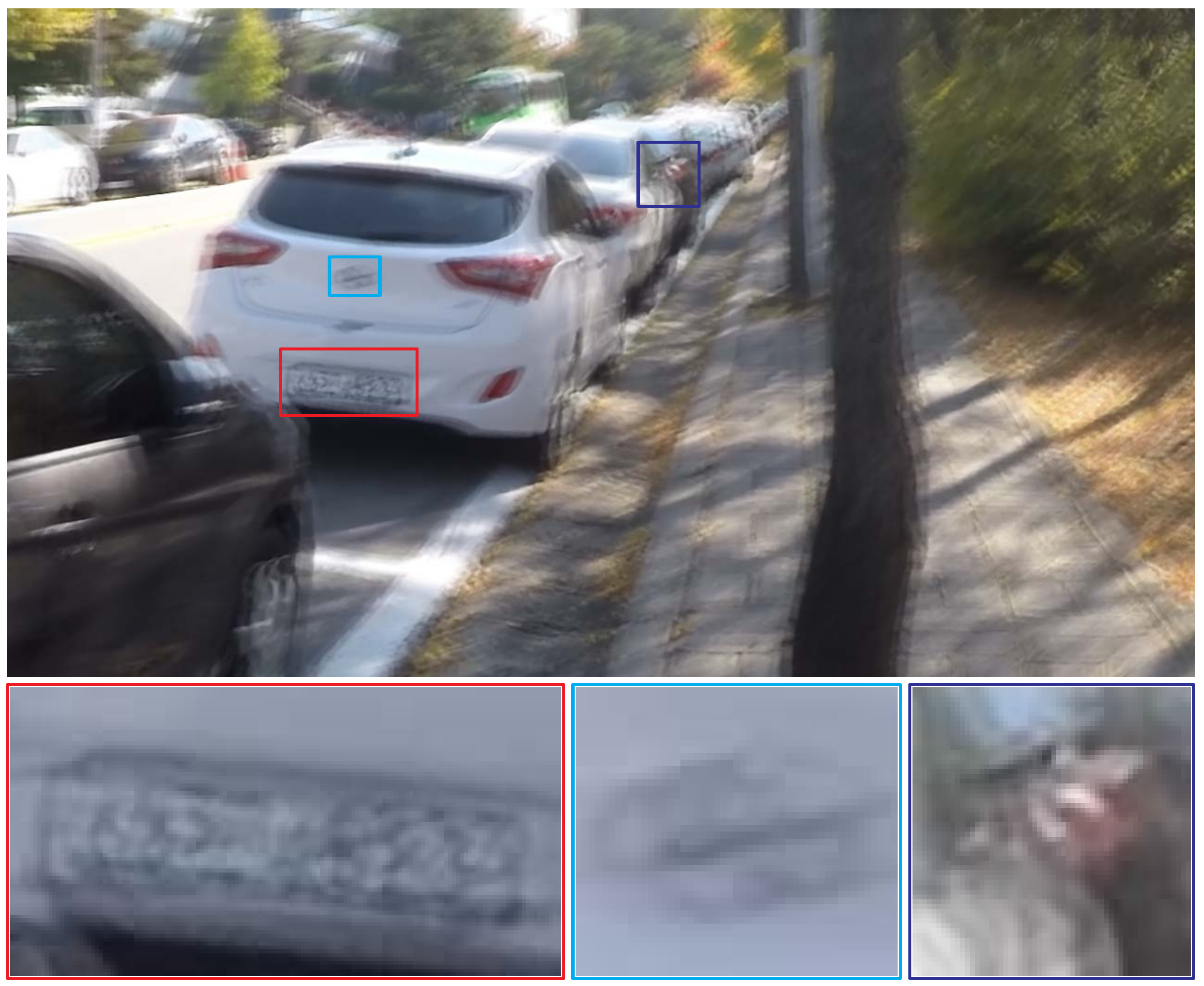} &\hspace{-4.5mm}
\includegraphics[width=0.245\linewidth]{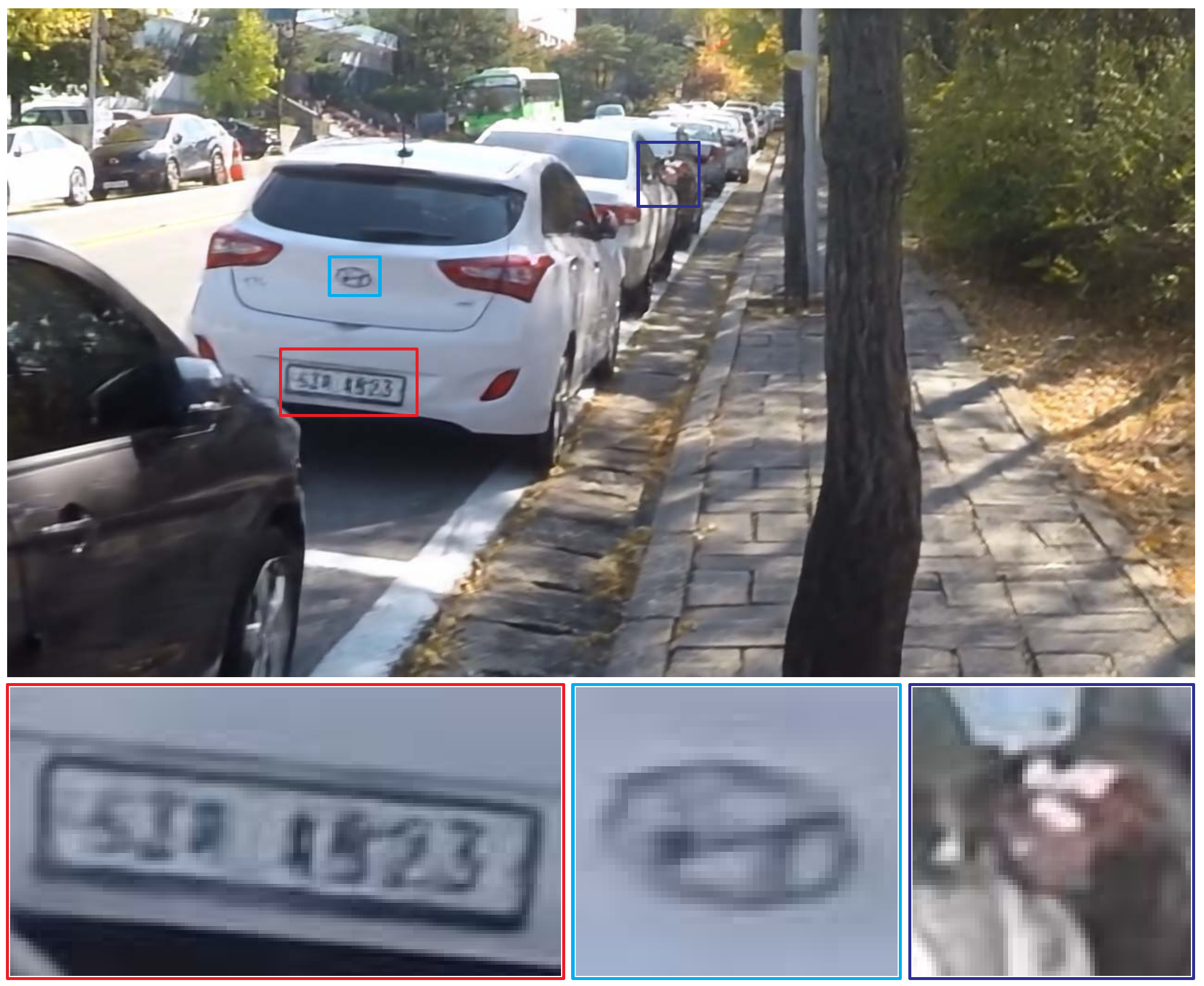} \\
(a) Blurred frame &\hspace{-4.5mm}  (b) GT &\hspace{-4.5mm}  (c) Kim and Lee~\cite{VD/kim/cvpr15}  &\hspace{-4.5mm}  (d) Tao et al.~\cite{SRN}\\
\includegraphics[width=0.245\linewidth]{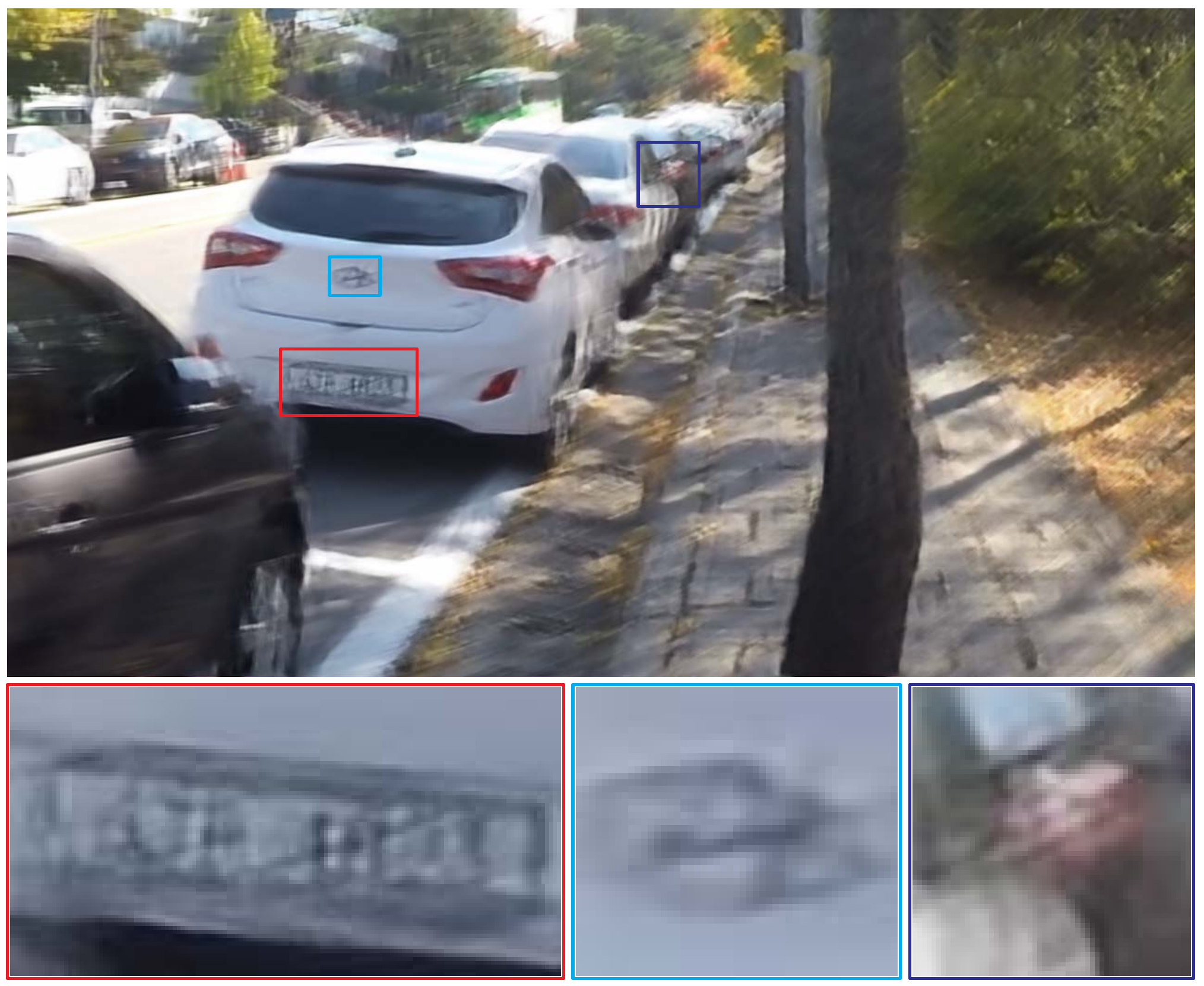}  &\hspace{-4.5mm}
\includegraphics[width=0.245\linewidth]{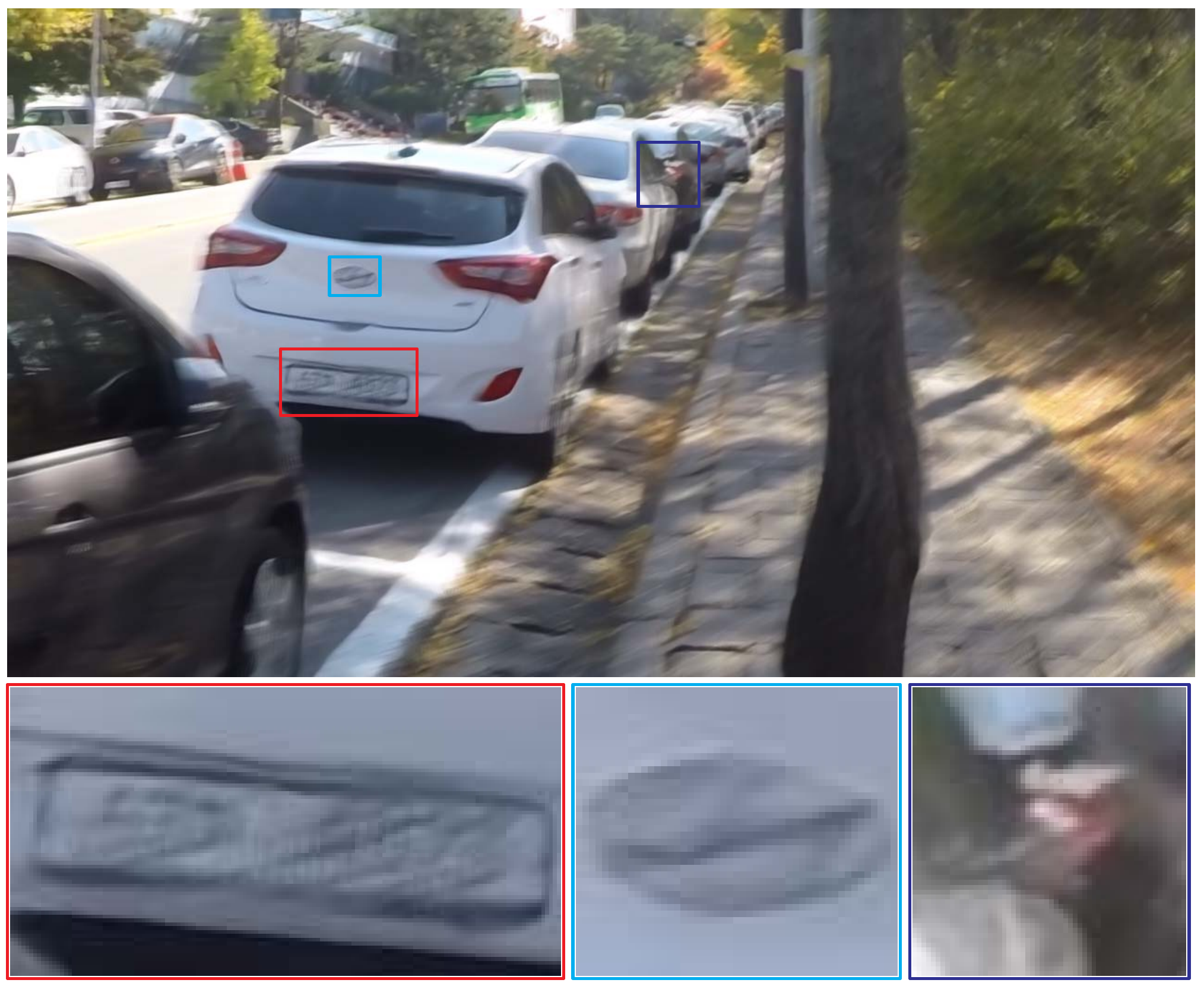} &\hspace{-4.5mm}
\includegraphics[width=0.245\linewidth]{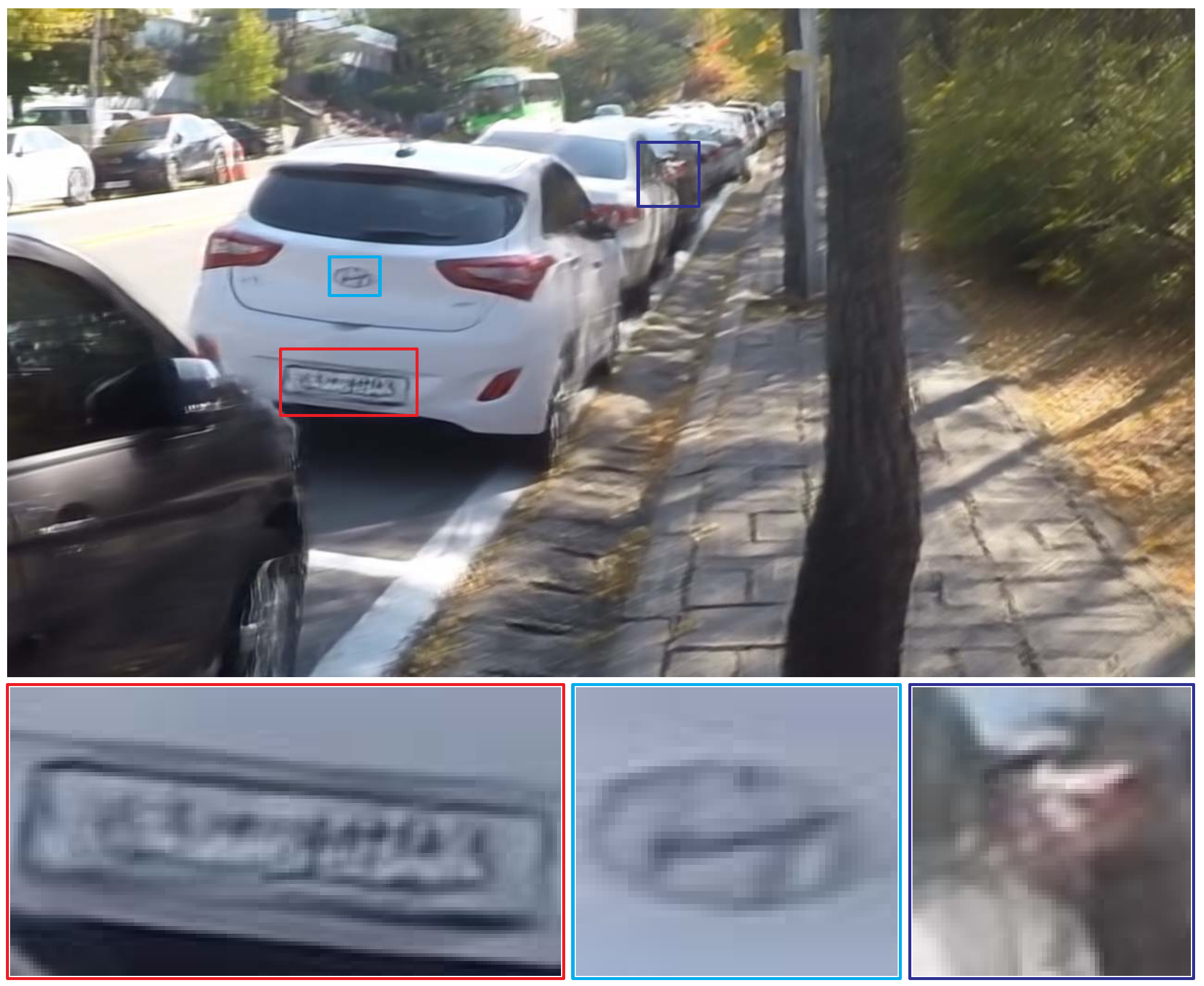} &\hspace{-4.5mm}
\includegraphics[width=0.245\linewidth]{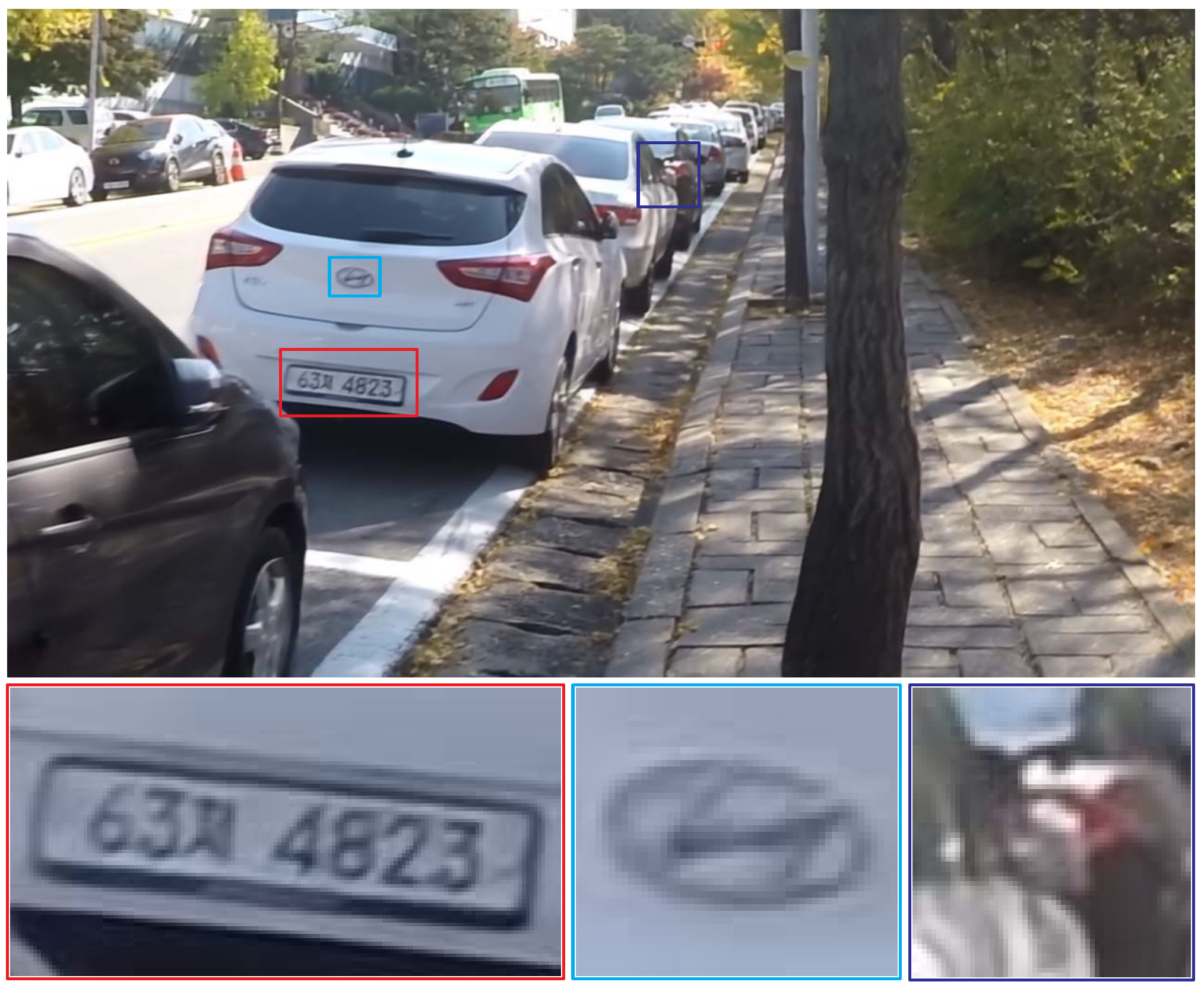} \\
(e) Su et al.~\cite{DVD/cvpr17} &\hspace{-4.5mm}  (f) EDVR~\cite{edvr} &\hspace{-4.5mm}  (g) STFAN~\cite{zhoushanchen/iccv19} &\hspace{-4.5mm}  (h) Ours \\
\end{tabular}
\caption{Deblurred results on the test dataset~\cite{MSCNN/deblur/cvpr17}. The proposed method generates much better deblurred images, where the license numbers are recognizable.}%
\label{fig: result-other-dataset-vis}
\vspace{-2mm}
\end{figure*}

Figure~\ref{fig: result-other-dataset-vis} shows some deblurred results from~\cite{MSCNN/deblur/cvpr17}.
We note that state-of-the-art methods do not generate sharp images and remove the non-uniform blur well.
In contrast, the proposed algorithm restores much clearer images, where the license numbers are recognizable.

\begin{figure*}[!t]\footnotesize
\centering
\begin{tabular}{ccccc}

\includegraphics[width=0.245\linewidth]{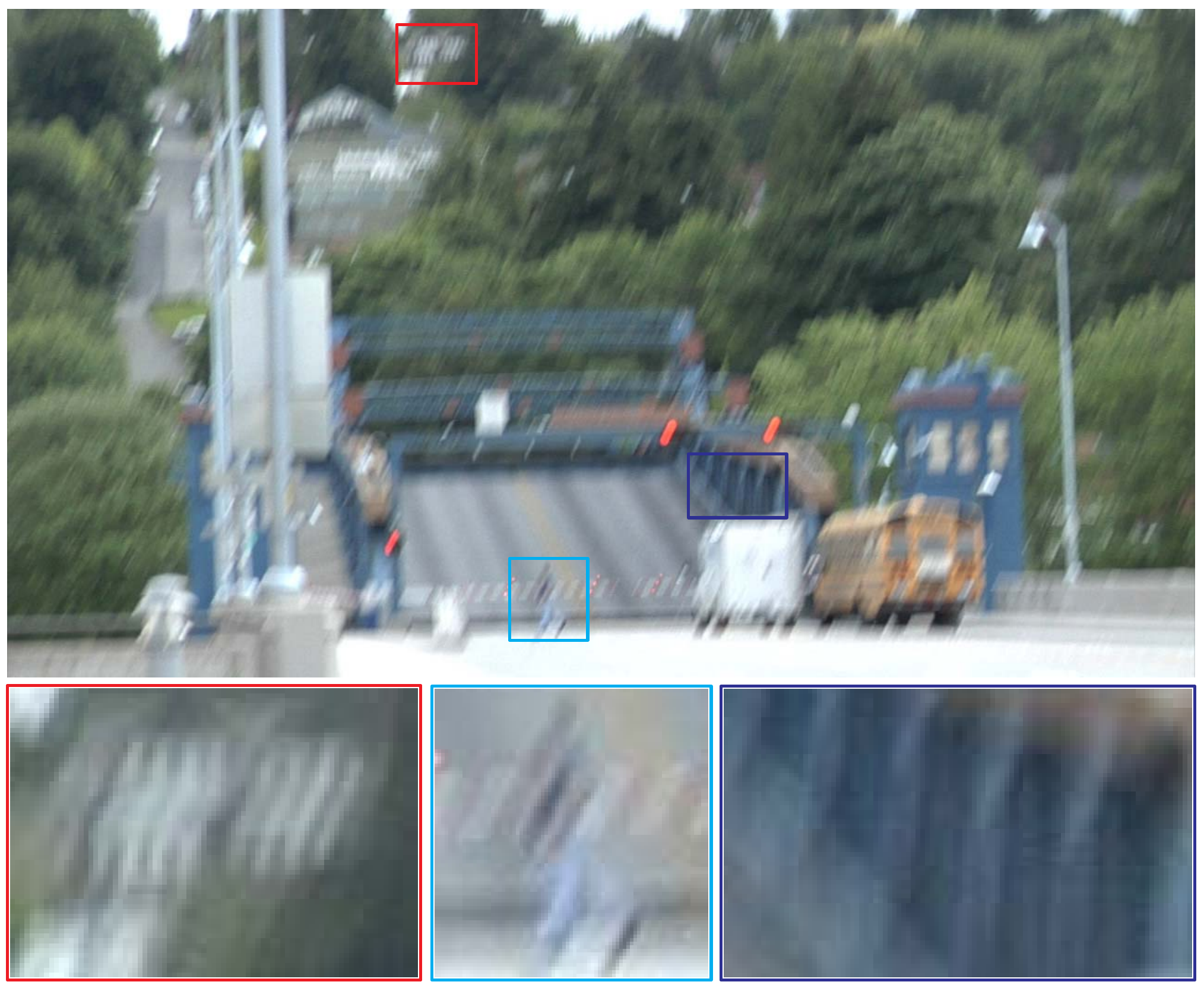} &\hspace{-4.5mm}
\includegraphics[width=0.245\linewidth]{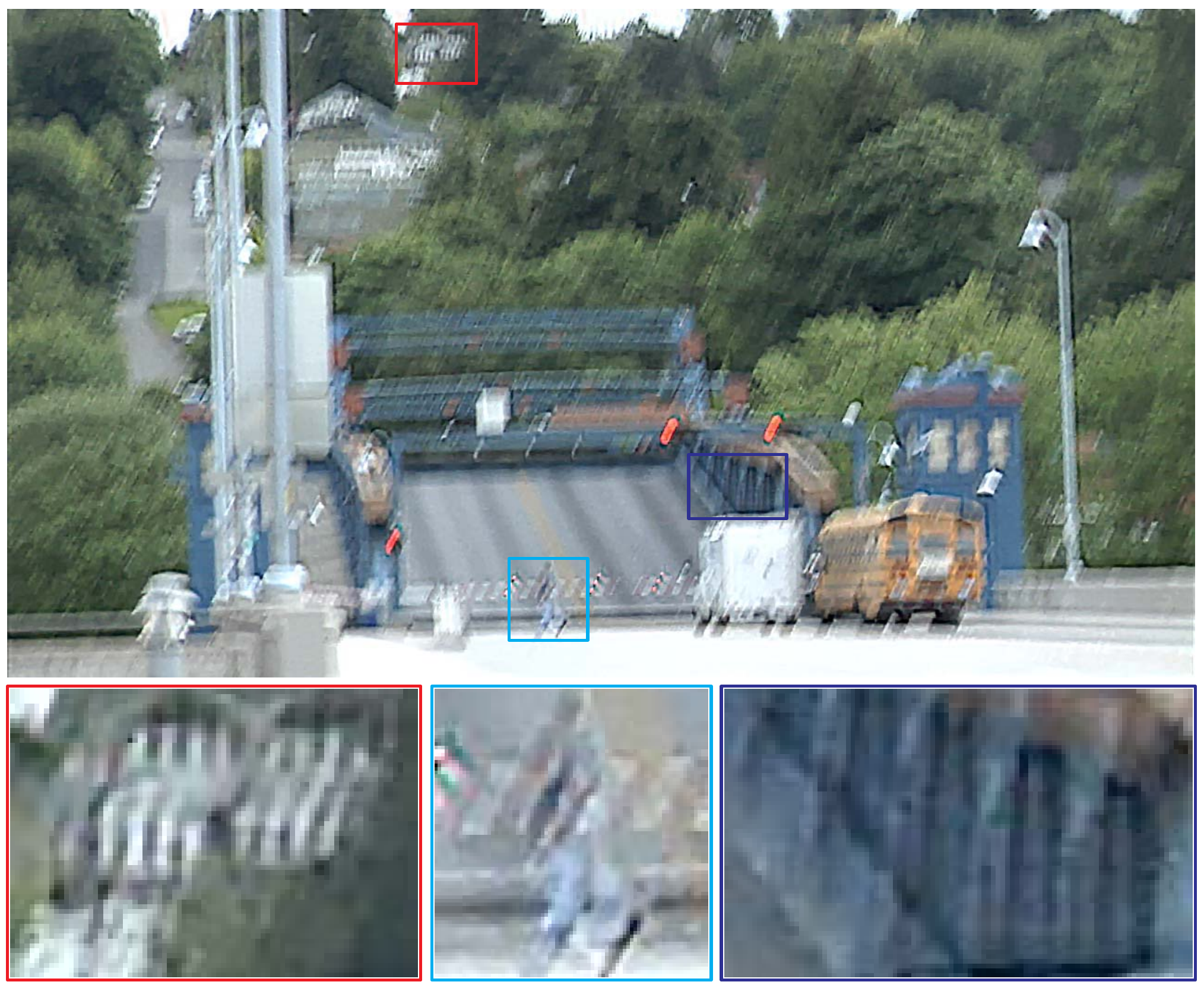} &\hspace{-4.5mm}
\includegraphics[width=0.245\linewidth]{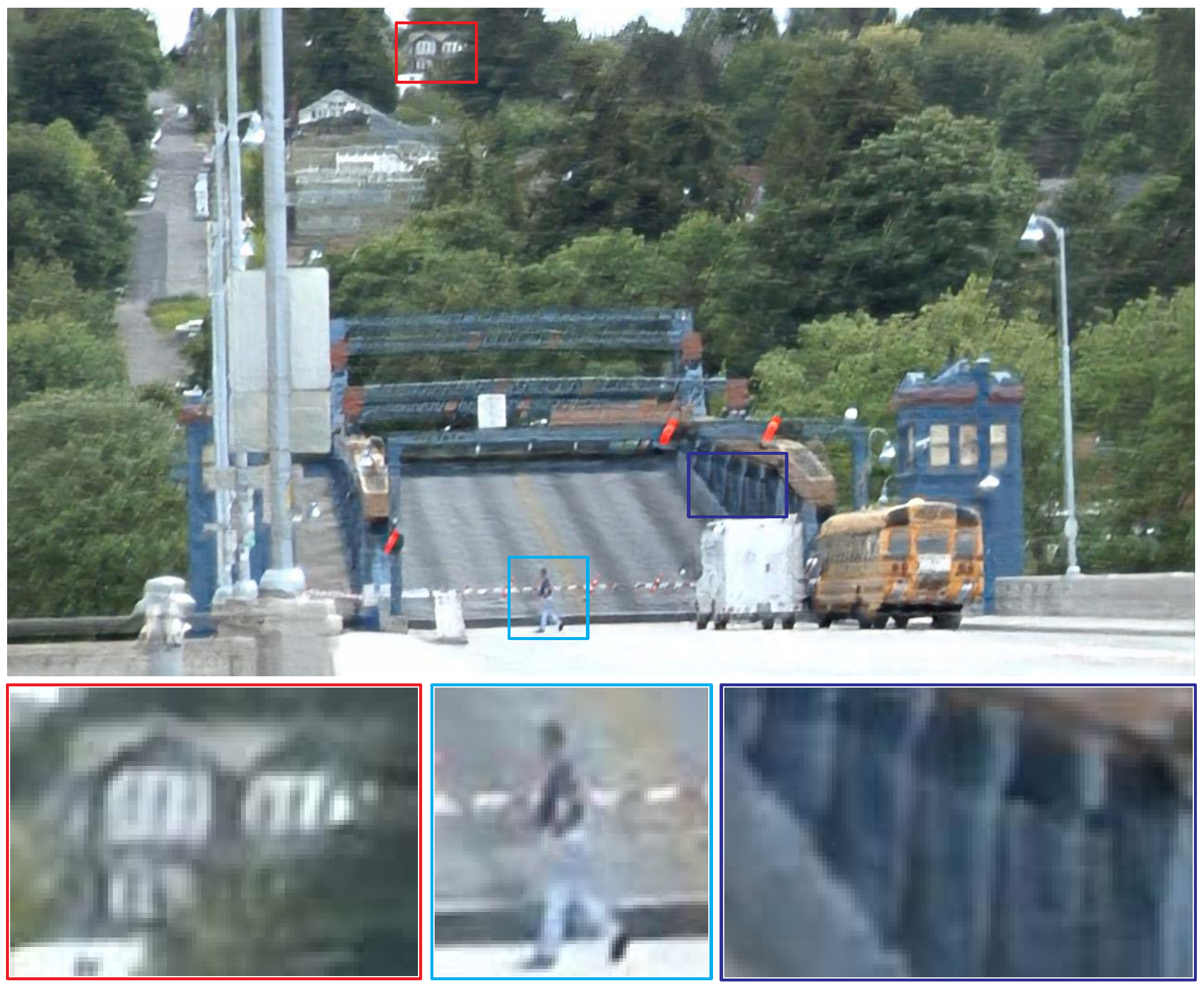} &\hspace{-4.5mm}
\includegraphics[width=0.245\linewidth]{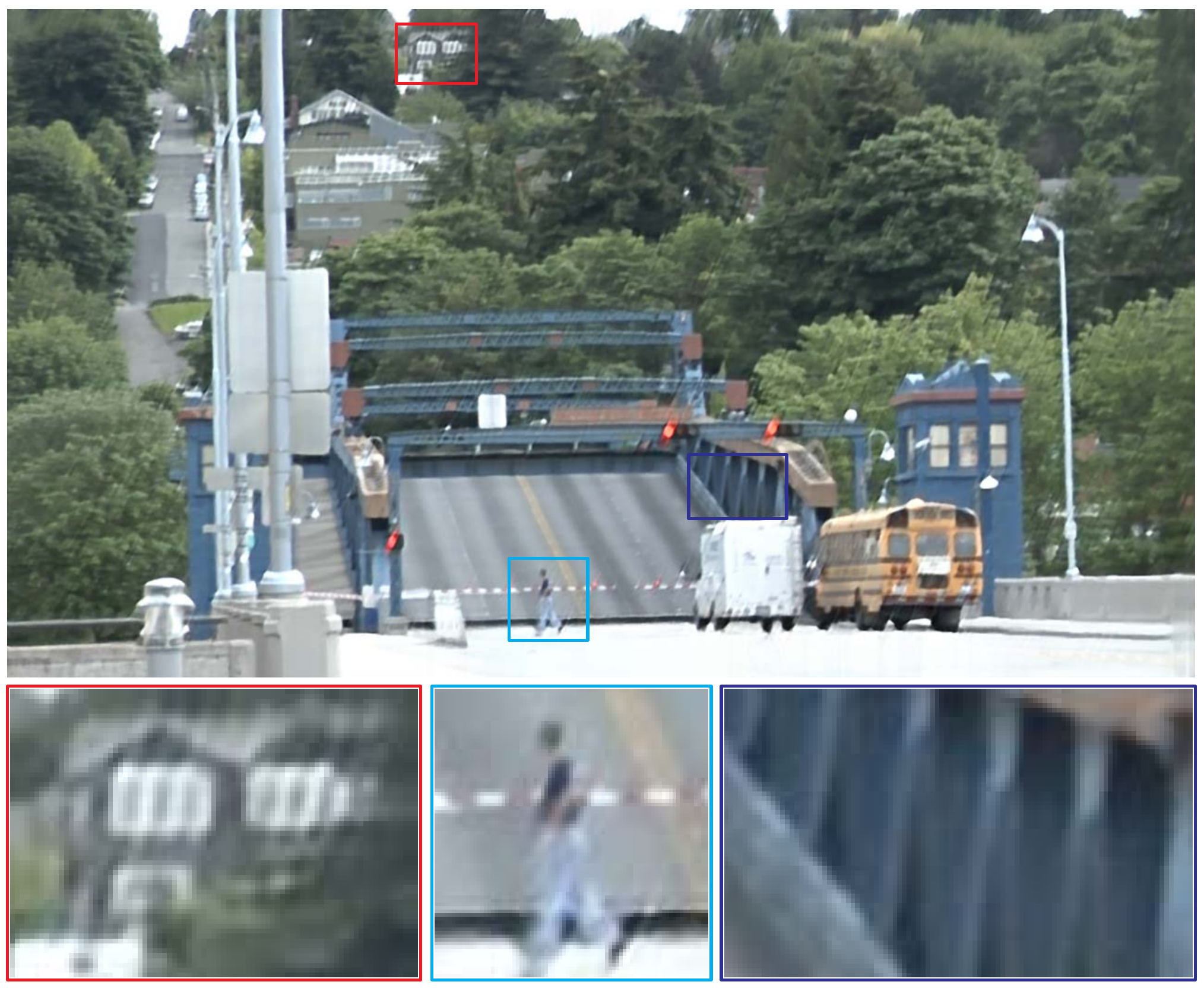} \\
(a) Blurred frame &\hspace{-4.5mm}  (b) Pan~et al.~\cite{darkchannel/tpami18} &\hspace{-4.5mm} (c) Cho et al.~\cite{cho/tog12/vd} &\hspace{-4.5mm}  (d) Kim and Lee~\cite{VD/kim/cvpr15}\\
\includegraphics[width=0.245\linewidth]{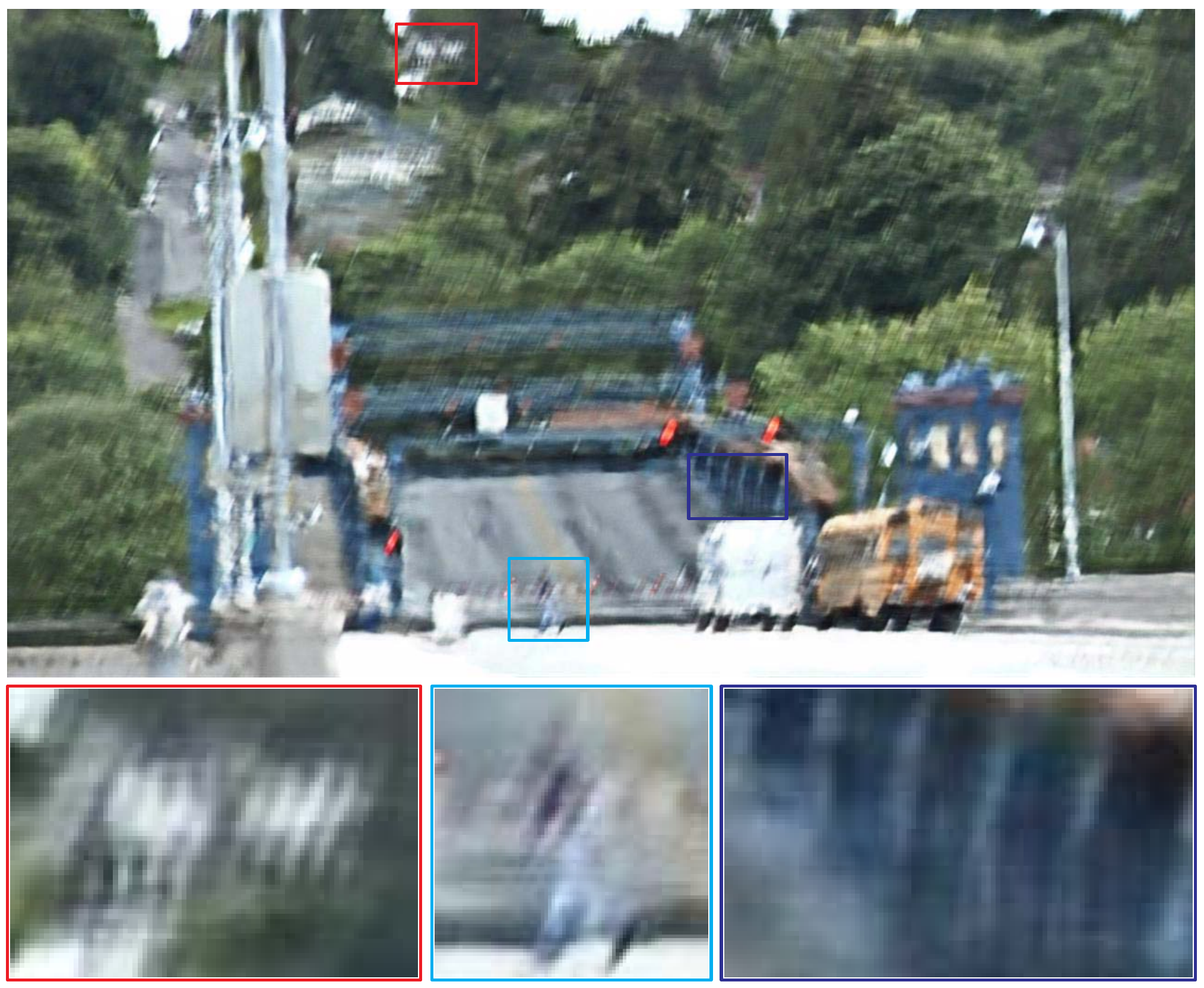} &\hspace{-4.5mm}
\includegraphics[width=0.245\linewidth]{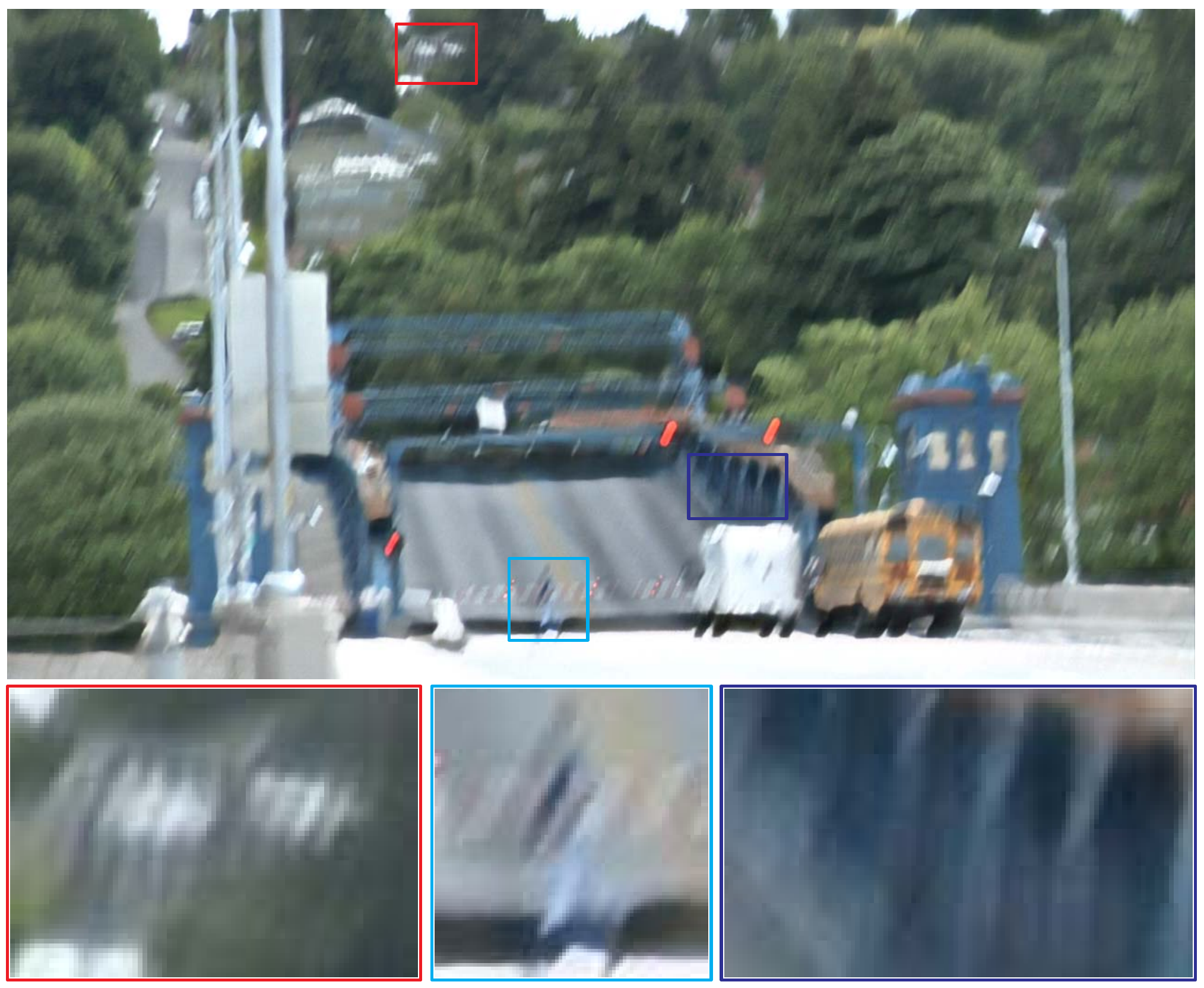} &\hspace{-4.5mm}
\includegraphics[width=0.245\linewidth]{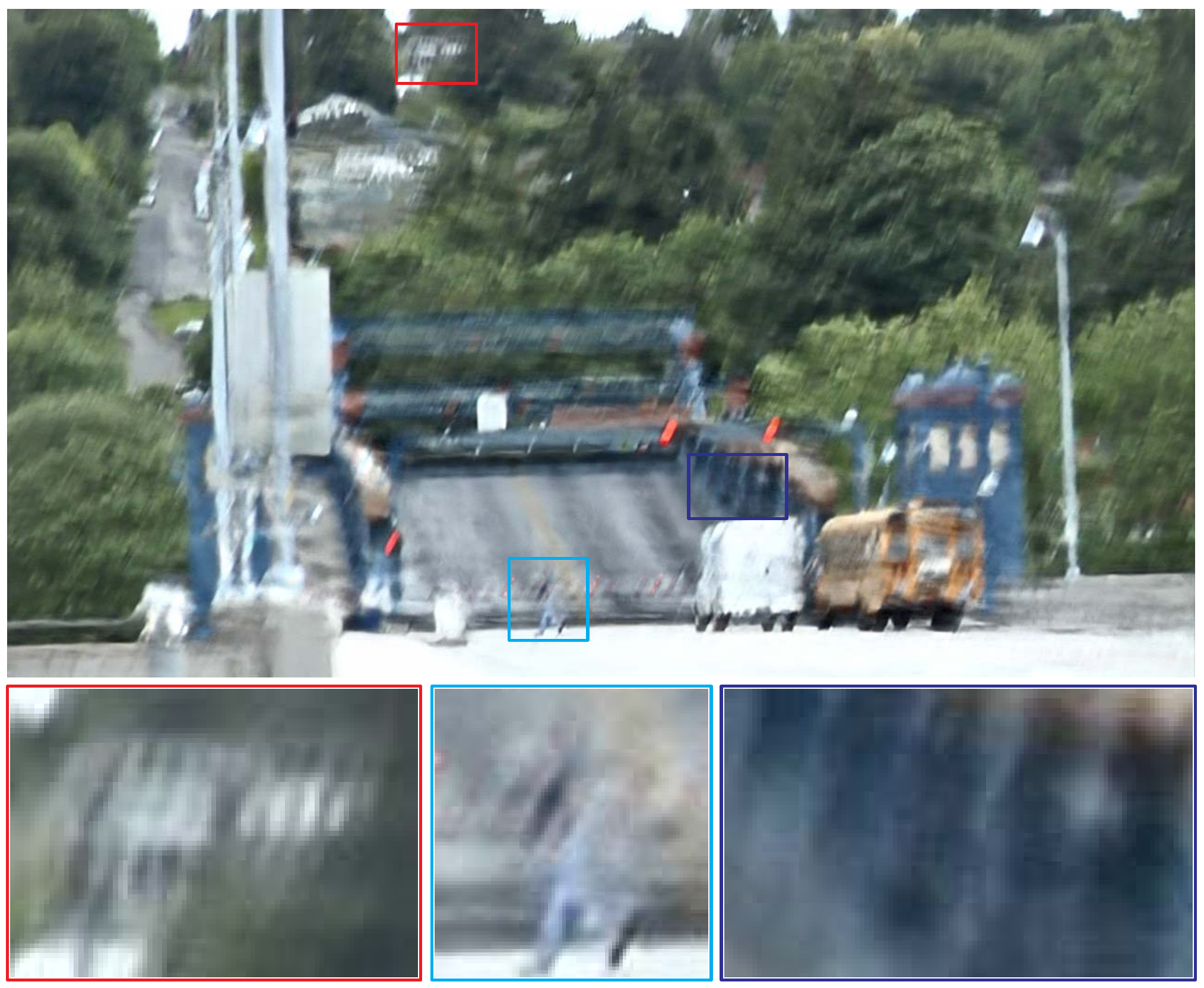} &\hspace{-4.5mm}
\includegraphics[width=0.245\linewidth]{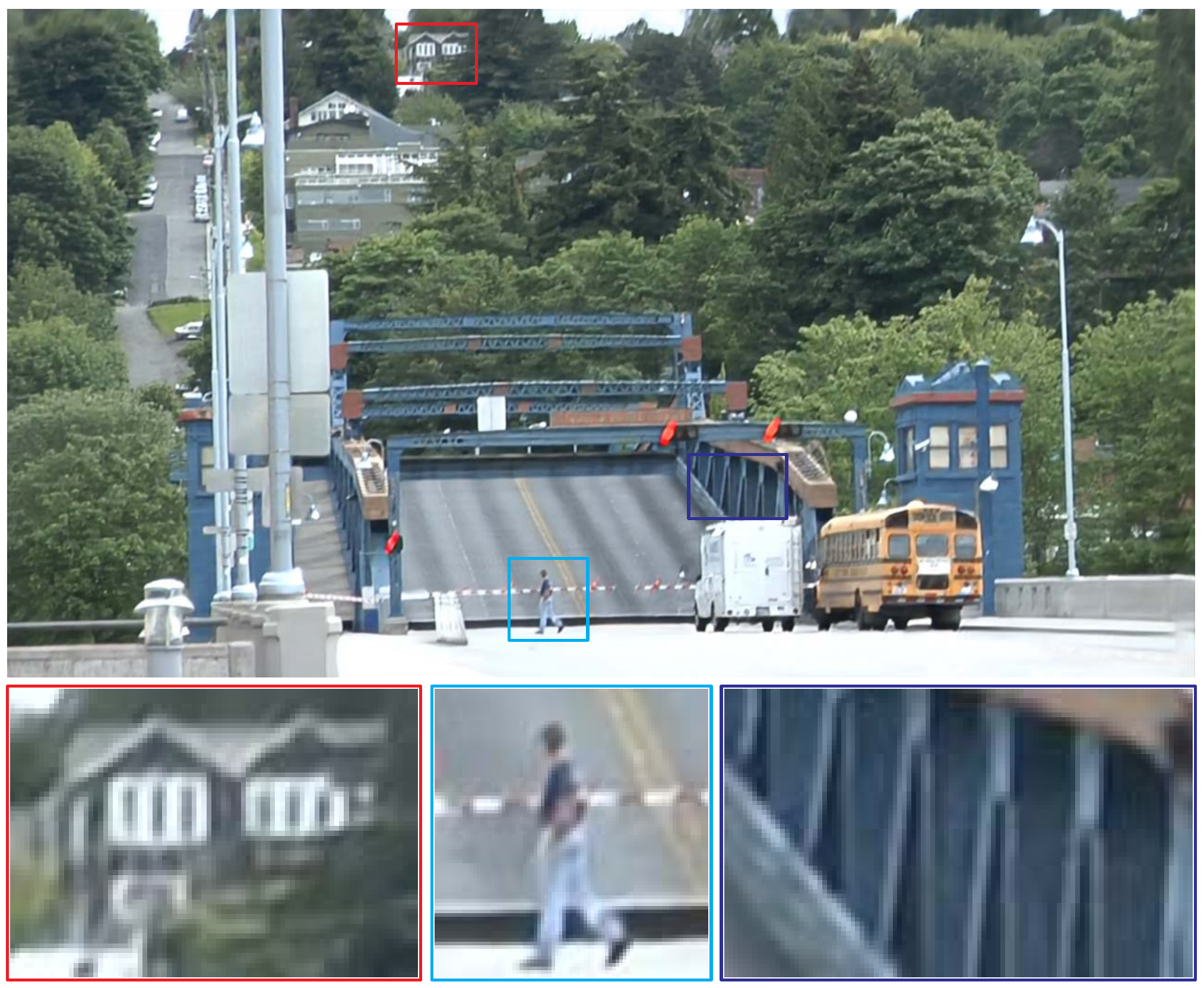}  \\
(e) Su et al.~\cite{DVD/cvpr17} &\hspace{-4.5mm}  (f) EDVR~\cite{edvr} &\hspace{-4.5mm}  (g) STFAN~\cite{zhoushanchen/iccv19}  &\hspace{-4.5mm} (h) Ours \\
\end{tabular}
\caption{Deblurred results on a real video from~\cite{cho/tog12/vd}. The proposed algorithm recovers a high-quality image with clearer details. }%
\label{fig: real-dataset-vis}
\vspace{-1mm}
\end{figure*}

\noindent{\bf Real examples.}
We further evaluate our algorithm on the real video deblurring dataset by Cho et al.~\cite{cho/tog12/vd}.
Figure~\ref{fig: real-dataset-vis} shows that the state-of-the-art methods~\cite{darkchannel/tpami18,cho/tog12/vd,VD/kim/cvpr15,zhoushanchen/iccv19,edvr,DVD/cvpr17} do not restore the sharp frames well.
Our algorithm generates much clearer frames with better detailed structures.
For example, the man and the boundaries of the buildings are much clearer (Figure~\ref{fig: real-dataset-vis}(h)).
%

\vspace{-2mm}
\section{Analysis and Discussions}
\vspace{-1mm}
We have shown that the proposed algorithm performs favorably against state-of-the-art methods. To better understand the proposed algorithm, we perform further analysis and discuss its limitations.
\vspace{-1mm}
\subsection{Effectiveness of the cascaded training}
\vspace{-1mm}
%
%
The proposed cascaded training algorithm ensures that the proposed method estimates optical flow from intermediate latent frames and updates the intermediate latent frames iteratively.
One may wonder whether the cascaded training algorithm helps video deblurring. To answer this question, we compare the method without using cascaded training algorithm (i.e., w/o CT in Table~\ref{tab: iteration-mask}), where we set stage number $T$ to be $1$ in Algorithm~\ref{alg: cascaded-training-algorithm} for fair comparisons.

\begin{figure}[!t]\footnotesize
\centering
\begin{tabular}{ccc}
\includegraphics[width=0.32\linewidth]{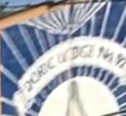} &\hspace{-4mm}
\includegraphics[width=0.32\linewidth]{figures/IMG_0037/00016_pwc_srn_baseline_crop} &\hspace{-4mm}
\includegraphics[width=0.32\linewidth]{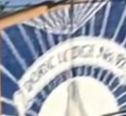} \\
(a) Blurred frame  &\hspace{-4mm}  (b) w/o CT &\hspace{-4mm}  (c) Stage 1\\
\includegraphics[width=0.32\linewidth]{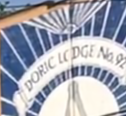} &\hspace{-4mm}
\includegraphics[width=0.32\linewidth]{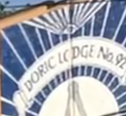} &\hspace{-4mm}
\includegraphics[width=0.32\linewidth]{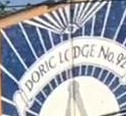}  \\
(d) Stage 2 &\hspace{-4mm}  (e) Stage 3&\hspace{-4mm}  (f) GT\\
\end{tabular}
\caption{Effectiveness of the cascaded training algorithm for video deblurring.
(b) denotes the deblurred result by the proposed method without using cascaded training.
(c)-(e) denote the results from stage 1, 2, and 3, respectively.
}%
\label{fig: cascaded-training}
\vspace{-5mm}
\end{figure}

Table~\ref{tab: iteration-mask} shows the quantitative evaluations on the benchmark dataset by Su et al.~\cite{DVD/cvpr17}.
We note that the method without using cascaded training algorithm estimates the optical flow from blurred inputs using PWC-Net, where this strategy is widely for image alignment in video deblurring~\cite{details/devil}.
However, this method does not generate high-quality deblurred results (Figure~\ref{fig: cascaded-training}(b)) as optical flow is related to the latent frames information instead of blurred ones during the exposure time.
In contrast, the proposed algorithm generates the results with higher PSNR and SSIM values.

We further compare the deblurred results generated by different stages in Table~\ref{tab: iteration-mask} and Figure~\ref{fig: cascaded-training}.
We note that using more stages generates better deblurred images. However, the improved performance is not significant.
Thus, we use two stages as a trade-off between accuracy and speed.
%
%
\begin{table}[!t]
  \caption{Effectiveness of the cascaded training algorithm for video deblurring, where ``CT" is the abbreviation of cascaded training.
  }
   \vspace{1mm}
   \label{tab: iteration-mask}
\footnotesize
 \centering
 \begin{tabular}{lcccc}
    \toprule
    Methods                   &w/o CT & Stage 1 & Stage 2 & Stage 3\\
    \hline
 PSNRs                &31.33          &31.59  & 32.13  & 32.20\\
 SSIMs                &0.9125         &0.9161 & 0.9268 & 0.9272\\
 \bottomrule
  \end{tabular}
\vspace{-1mm}
\end{table}

\begin{figure}[!t]\footnotesize
\centering
\begin{tabular}{cccc}
\includegraphics[width=0.32\linewidth]{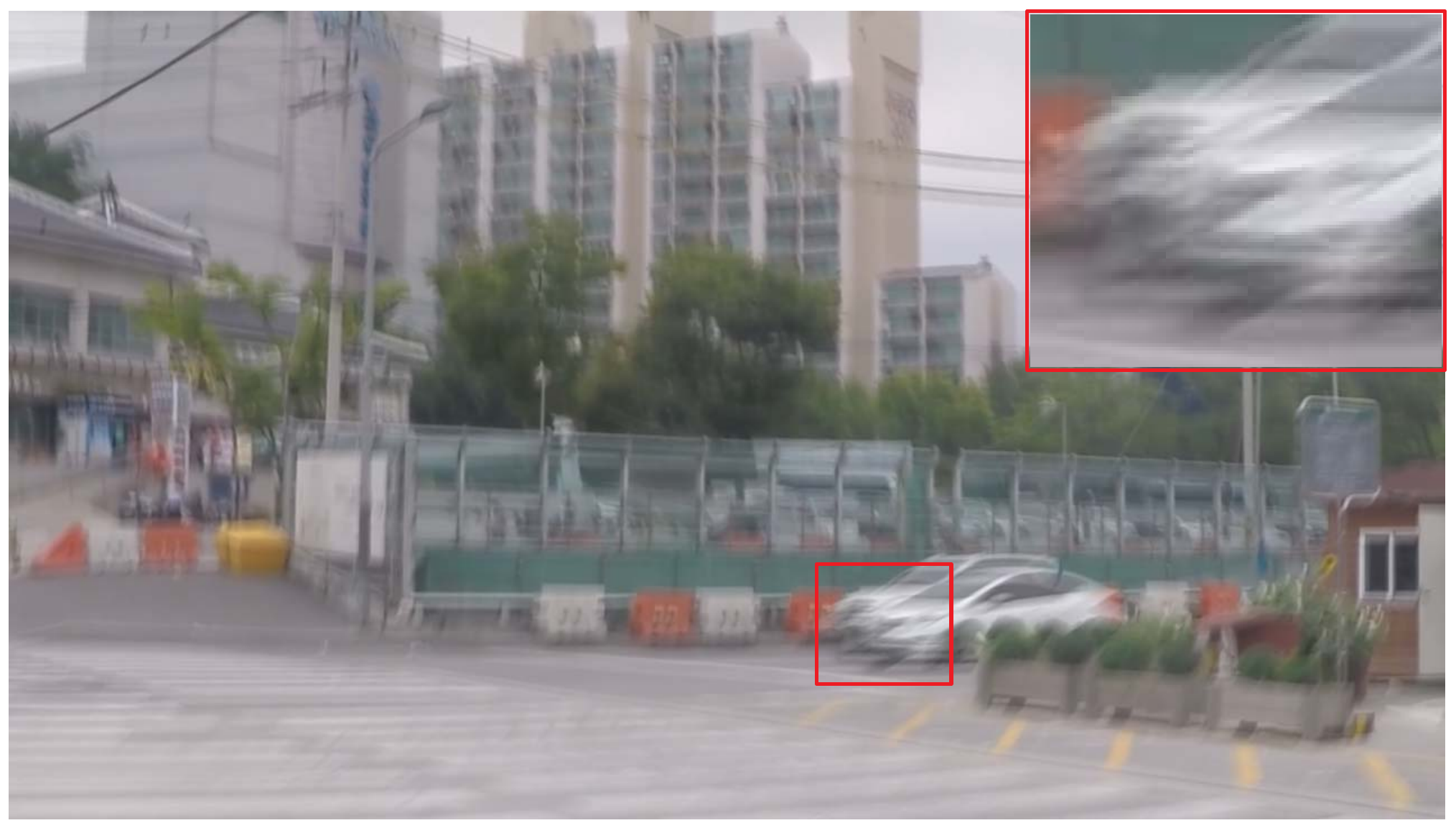} &\hspace{-4.5mm}
\includegraphics[width=0.32\linewidth]{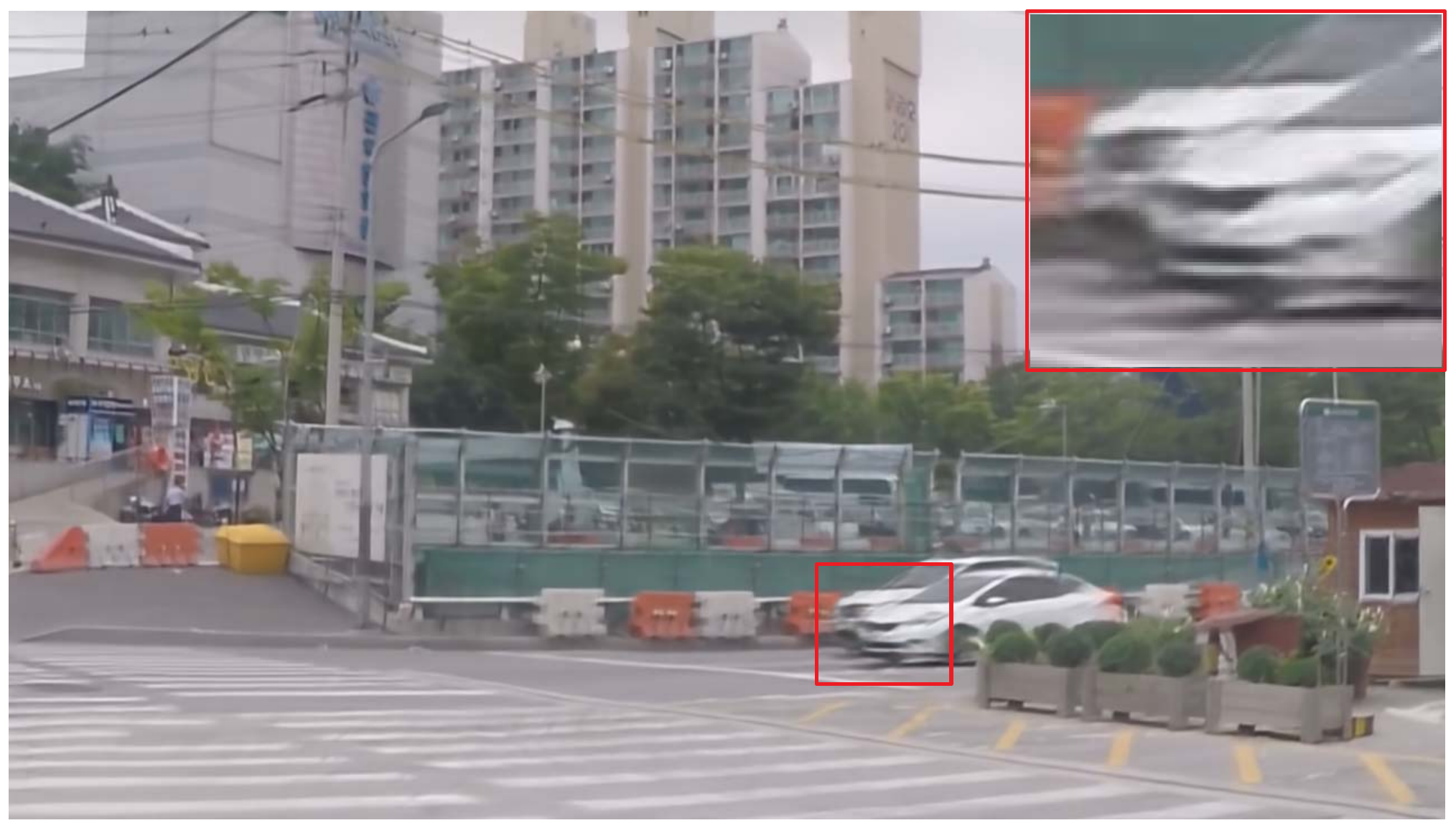} &\hspace{-4.5mm}
\includegraphics[width=0.32\linewidth]{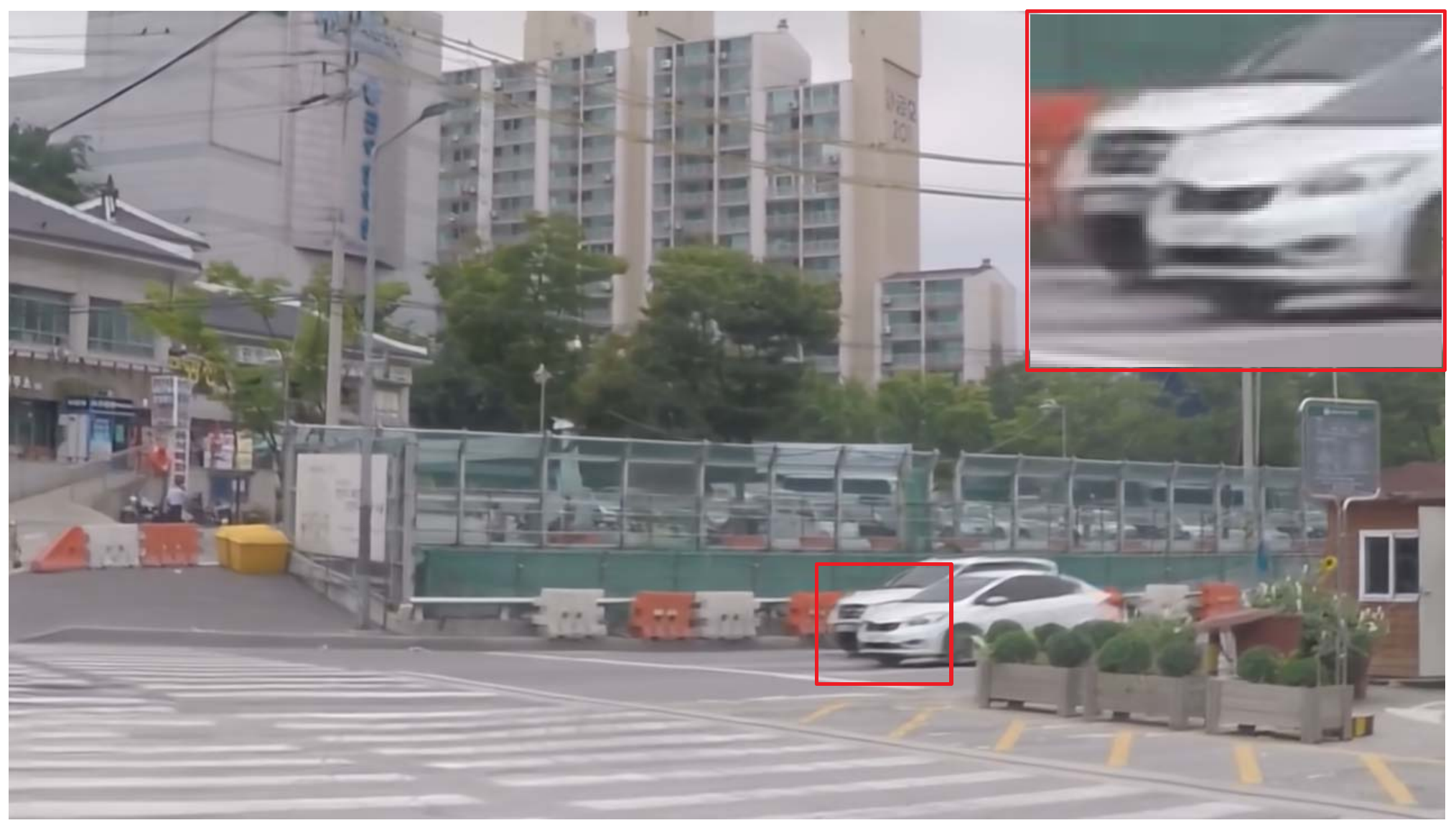} \\
 (a) Blurred frame &\hspace{-4mm}  (b) w/o CT &\hspace{-4mm}  (c) Ours\\
\includegraphics[width=0.32\linewidth]{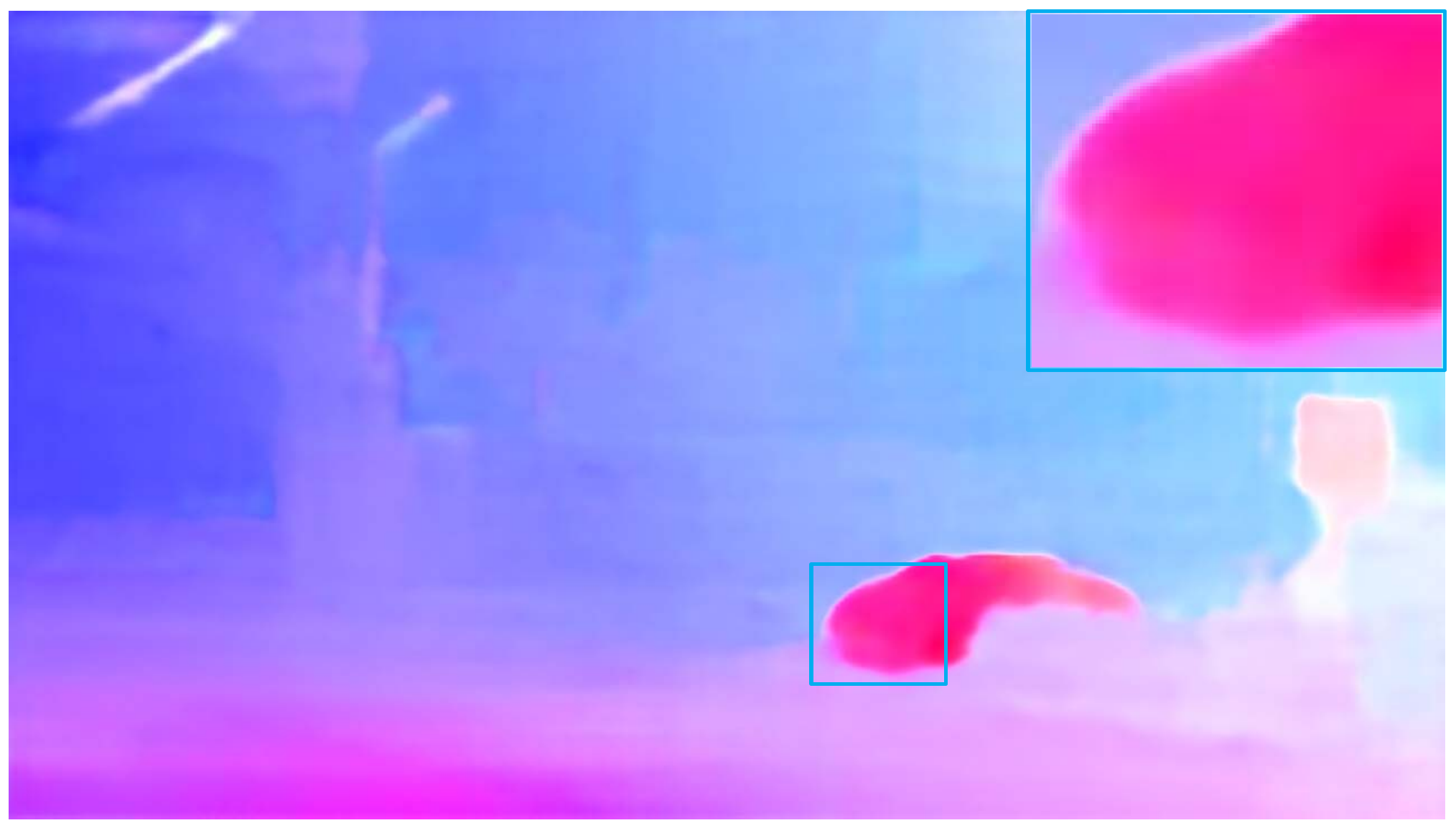}  &\hspace{-4.5mm}
\includegraphics[width=0.32\linewidth]{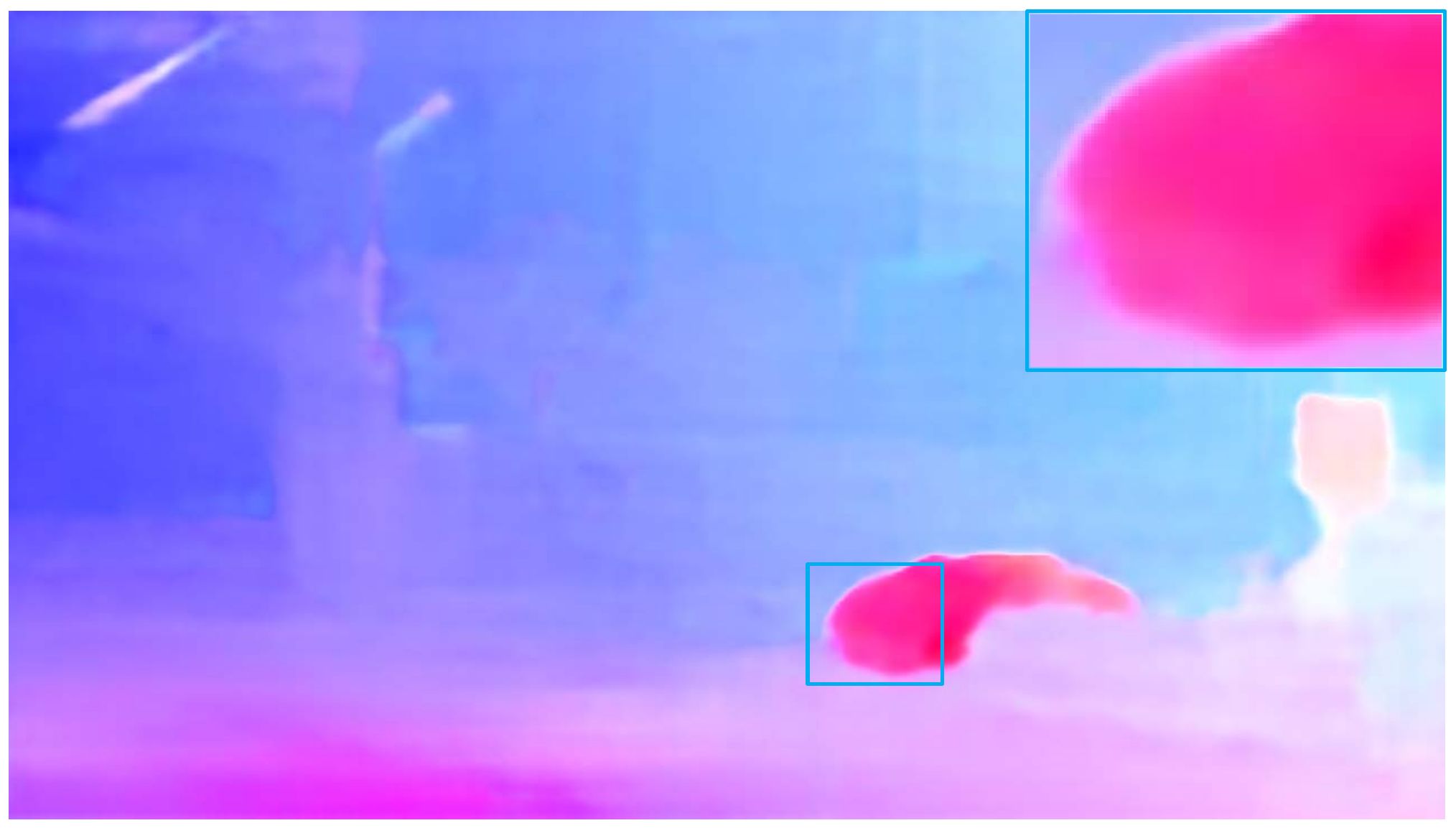}  &\hspace{-4.5mm}
\includegraphics[width=0.32\linewidth]{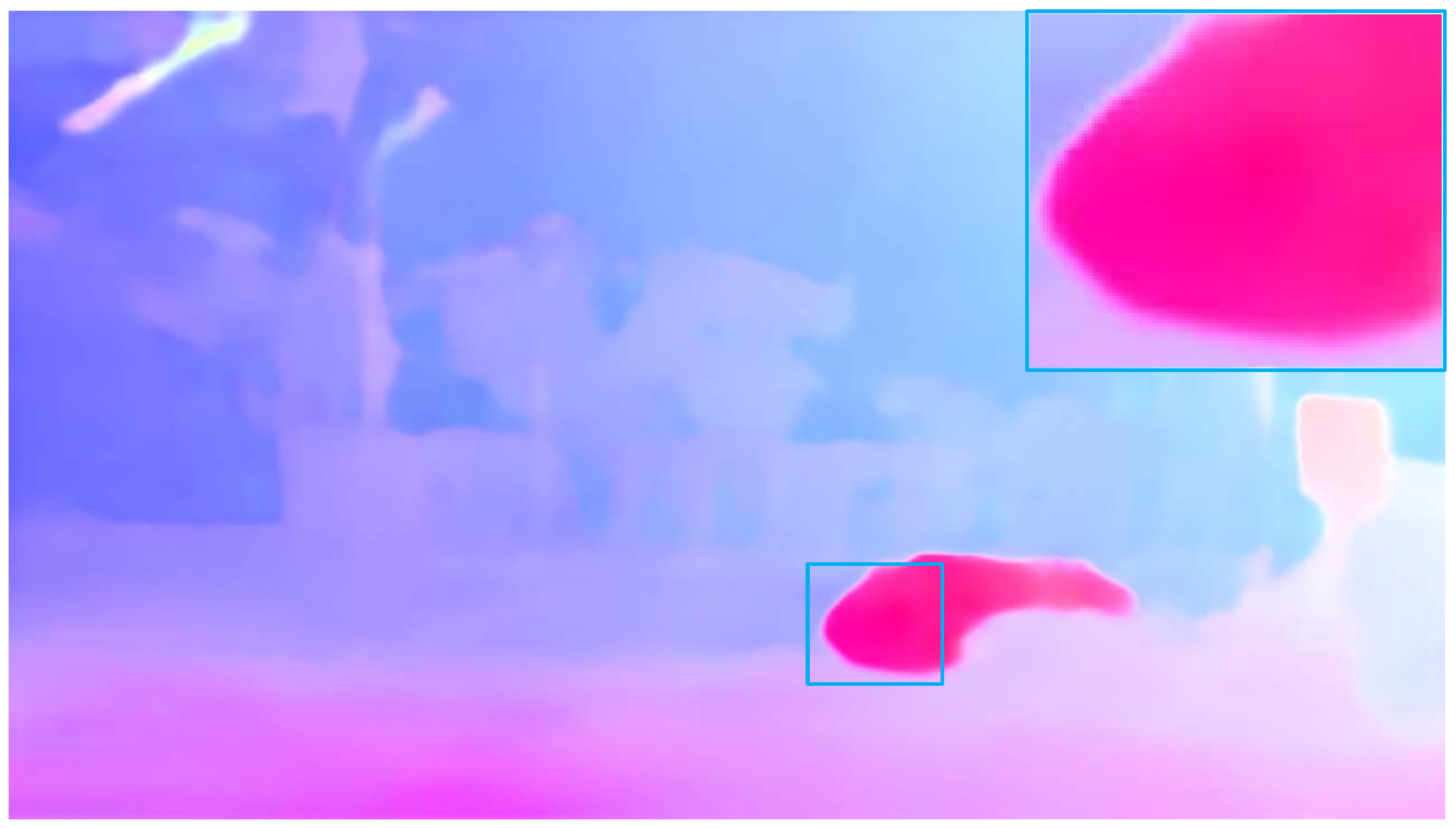} \\
\hspace{-4mm} (d) w/o CT &\hspace{-4mm}  (e) Stage 1 &\hspace{-4mm}  (f) Stage 2 \\
\end{tabular}
\caption{Effect of optical flow on video deblurring. The optical flow by the proposed method contains sharp boundaries well (see (f)), which facilitates the latent frame restoration.}%
\label{fig: effect-optical-flow}
\vspace{-1mm}
\end{figure}

We further note that directly estimating optical flow from blurred inputs will increase ambiguity at frame boundaries for video deblurring.
Figure~\ref{fig: effect-optical-flow}(d) demonstrates that the boundaries of the estimated optical flow are blurry, which accordingly affects the important boundaries restoration (Figure~\ref{fig: effect-optical-flow}(b)).
In contrast, the optical flow by the proposed method contains sharp boundaries well (Figure~\ref{fig: effect-optical-flow}(f)), which facilitates the latent frame restoration (Figure~\ref{fig: effect-optical-flow}(c)).

\subsection{Effectiveness of the temporal sharpness prior}
\begin{figure}[!t]\footnotesize
\centering
\begin{tabular}{cccc}
%
\hspace{-4mm}
\includegraphics[width=0.24\linewidth]{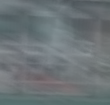} &\hspace{-4mm}
\includegraphics[width=0.24\linewidth]{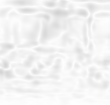} &\hspace{-4mm}
\includegraphics[width=0.24\linewidth]{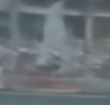} &\hspace{-4mm}
\includegraphics[width=0.24\linewidth]{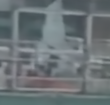} \\
\hspace{-4mm}
\includegraphics[width=0.24\linewidth]{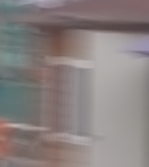} &\hspace{-4mm}
\includegraphics[width=0.24\linewidth]{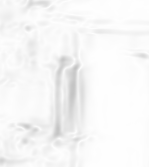} &\hspace{-4mm}
\includegraphics[width=0.24\linewidth]{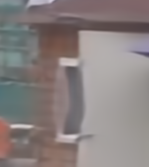} &\hspace{-4mm}
\includegraphics[width=0.24\linewidth]{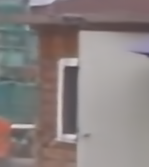} \\
\hspace{-4mm} (a)  &\hspace{-4mm}  (b) &\hspace{-4mm}  (c) &\hspace{-4mm}  (d) \\
\end{tabular}
\caption{Effectiveness of the temporal sharpness prior.
(a) Blurred input.
(b) Visualizations of the intermediate temporal sharpness prior.
(c)-(d) denote the results without and with the temporal sharpness prior, respectively. }%
\label{fig: sharp-prior}
\end{figure}

We develop a temporal sharpness prior to better explore the properties of consecutive frames so that it makes the deep CNN models more compact.
To demonstrate the effectiveness of this prior, we disable this prior in the proposed method and retrain the algorithm without using the temporal sharpness prior with the same settings for fair comparisons.
We evaluate the temporal sharpness prior on 4 videos with significant blur effects from the test dataset~\cite{DVD/cvpr17}.
Table~\ref{tab: sharp-prior} and Figure~\ref{fig: sharp-prior} show both quantitative and qualitative evaluations.
We note that the temporal sharpness prior is able to distinguish the sharpness pixels and blurred pixels from adjacent frames so that it can help the deep CNN model for better frame restoration.
Figure~\ref{fig: sharp-prior}(b) shows the visualizations of the temporal sharpness prior, where the blurred pixels can be better detected.
The comparisons in Table~\ref{tab: sharp-prior} demonstrate that using the temporal sharpness prior is able to improve the accuracy of video deblurring.
Figure~\ref{fig: sharp-prior} further shows that using the temporal sharpness prior is able to generate the frames with clearer structures.
%
\begin{table}[!t]
  \caption{Effectiveness of the temporal sharpness prior on video deblurring.
  }
   \vspace{1mm}
   \label{tab: sharp-prior}
\footnotesize
 \centering
 \begin{tabular}{lcc}
    \toprule
    Methods                   &w/o temporal sharpness prior & Ours\\
    \hline
 PSNRs                &34.48         &\bf{34.63}\\
 SSIMs                &0.9126         &\bf{0.9268}\\
 \bottomrule
  \end{tabular}
\end{table}

\begin{table}[!t]
  \caption{Effect of the optical flow estimation module.
  }
   \vspace{1mm}
   \label{tab: optical-flow-effect}
\footnotesize
 \centering
 \begin{tabular}{lccc}
    \toprule
    Methods              &w/o optical flow  &w/ FlowNet 2.0    &w/ PWC-Net\\
    \hline
 PSNRs           &31.19    &32.06         &32.13         \\
 SSIMs           &0.9055   &0.9254       &0.9268         \\
 \bottomrule
  \end{tabular}
\vspace{-4mm}
\end{table}

%
%

\subsection{Effect of optical flow}
As several algorithms either directly concatenate consecutive frames~\cite{DVD/cvpr17} or estimate filter kernels~\cite{zhoushanchen/iccv19} instead of using optical flow for video deblurring,
one may wonder whether optical flow helps video deblurring.
To answer this question, we remove the optical flow estimation module and compare with the method that directly concatenates consecutive frames as the input of the restoration network $\mathcal{N}_l$\footnote{The proposed method without using optical flow reduces to the network $\mathcal{N}_l$ which takes the concatenation of consecutive frames as the input.}.
Table~\ref{tab: optical-flow-effect} shows that using optical flow is able to improve the performance of video deblurring.

In addition, we further evaluate the optical flow estimation module using FlowNet 2.0~\cite{FlowNet2/cvpr17}.
Table~\ref{tab: optical-flow-effect} shows that the proposed method is robust to optical flow modules.

\vspace{-1mm}
\subsection{Model size}
\vspace{-1mm}
As stated in Section~\ref{sec: introduction}, we aim to improve the accuracy of video deblurring while do not increase the model capacity using domain knowledge of video deblurring.
Table~\ref{tab: run-time} shows that the proposed algorithm has a relatively smaller model size against state-of-the-art methods.
Compared to the baseline models, the proposed model does not increase any model size while generating much better results.

%
\begin{table}[!t]\footnotesize
  \caption{Comparisons of model sizes against state-of-the-art methods and baselines. ``TSP" is the abbreviation of temporal sharpness prior.
  }
   \vspace{1mm}
   \label{tab: run-time}
\resizebox{0.49\textwidth}{!}{
 \centering
 \begin{tabular}{ccccccccccc}
    \toprule
    Methods                  &Su et al.~\cite{DVD/cvpr17}               &EDVR~\cite{edvr}           &w/o CT    & w/o TSP     &Ours \\
    \hline
 Model size                 &15.30M             &23.60M            &16.19M      &16.19M     &16.19M\\
 \bottomrule
  \end{tabular}
}
\vspace{-1mm}
\end{table}
\subsection{Limitations}
%
Although the temporal sharpness prior is effective for videos with significant blur, it is less effective when the blur exists in each position of all frames.
In such cases, the temporal sharpness prior is less likely to distinguish whether the pixel is clearer or not.
%
%
Table~\ref{tab: sharp-prior-limitation} shows the deblurred results on $3$ videos from the test dataset~\cite{DVD/cvpr17}, where each position in a frame contains blur effect.
We note that using the temporal sharpness prior does not improve the deblurring performance significantly.
%
%

\begin{table}[!t]
  \caption{Evaluations on the blurred videos, where the blur exists in each position of each frame. The temporal sharpness prior is less effective when it fails to identify the clear pixels.
  }
   \vspace{1mm}
   \label{tab: sharp-prior-limitation}
\footnotesize
 \centering
 \begin{tabular}{lcc}
    \toprule
    Methods                   &w/o temporal sharpness prior & Ours\\
    \hline
 PSNRs                &31.31         &31.33\\
 SSIMs                &0.9238         &0.9239\\
 \bottomrule
  \end{tabular}
\vspace{-4mm}
\end{table}
\vspace{-1mm}
\section{Concluding Remarks}
\vspace{-1mm}
We have proposed a simple and effective deep CNN model for video deblurring.
The proposed CNN explores the simple and well-established principles used in the variational model-based methods and mainly consists of optical flow estimation from intermediate latent frames and latent frame restoration.
We have developed a temporal sharpness prior to help the latent image restoration and an effective cascaded training approach to train the proposed CNN model.
By training in an end-to-end manner, we have shown that the proposed CNN model is more compact and efficient and performs favorably against state-of-the-art methods on both benchmark datasets and real-world videos.

%

\section{Supplemental Material}
\label{sec: Network Details}
\begin{figure*}[h]\footnotesize
\centering
\begin{tabular}{c}
\includegraphics[width = 0.98\linewidth]{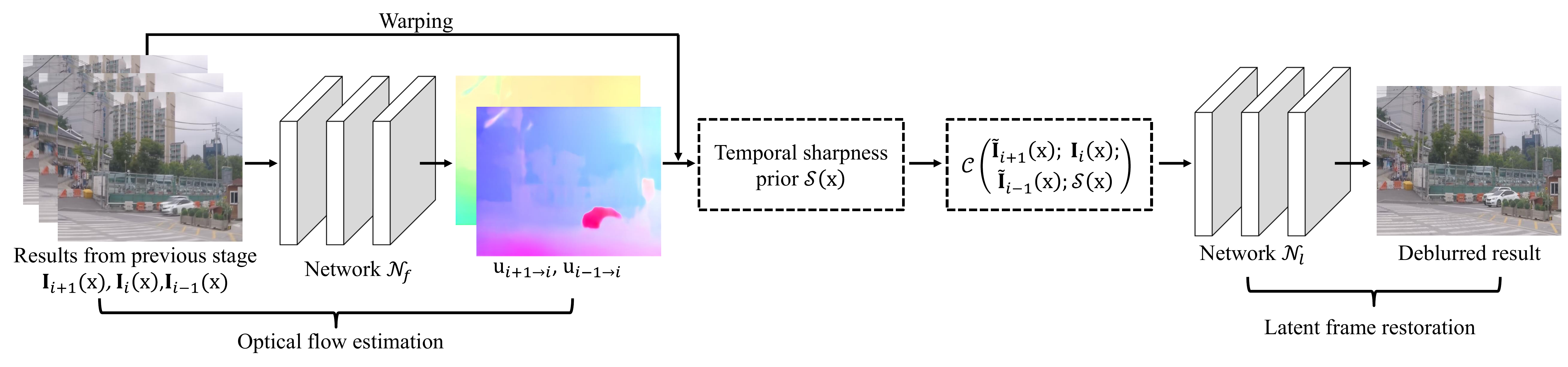}\\
\end{tabular}
\caption{An overview of the proposed method at one stage. The proposed algorithm contains the optical flow estimation module, latent image restoration module, and the temporal sharpness prior.
All the modules are jointly trained in an end-to-end manner.
At each stage, it takes three adjacent frames estimated from the previous stage as the input and generates the deblurred results of the central frame.
When handling every three adjacent frames, the proposed network shares the same network parameters.
The variables $\tilde{\mathbf{I}}_{i+1}(\mathrm{x})$ and $\tilde{\mathbf{I}}_{i-1}(\mathrm{x})$ denote the warped results of ${\mathbf{I}}_{i+1}(\mathrm{x} + \mathbf{u}_{i+1\to i})$ and ${\mathbf{I}}_{i-1}(\mathrm{x} + \mathbf{u}_{i-1\to i})$, respectively.
}
\label{fig: flow-chart}
%
\end{figure*}

\begin{figure*}[h]\footnotesize
\centering
\begin{tabular}{c}
\includegraphics[width = 0.76\linewidth]{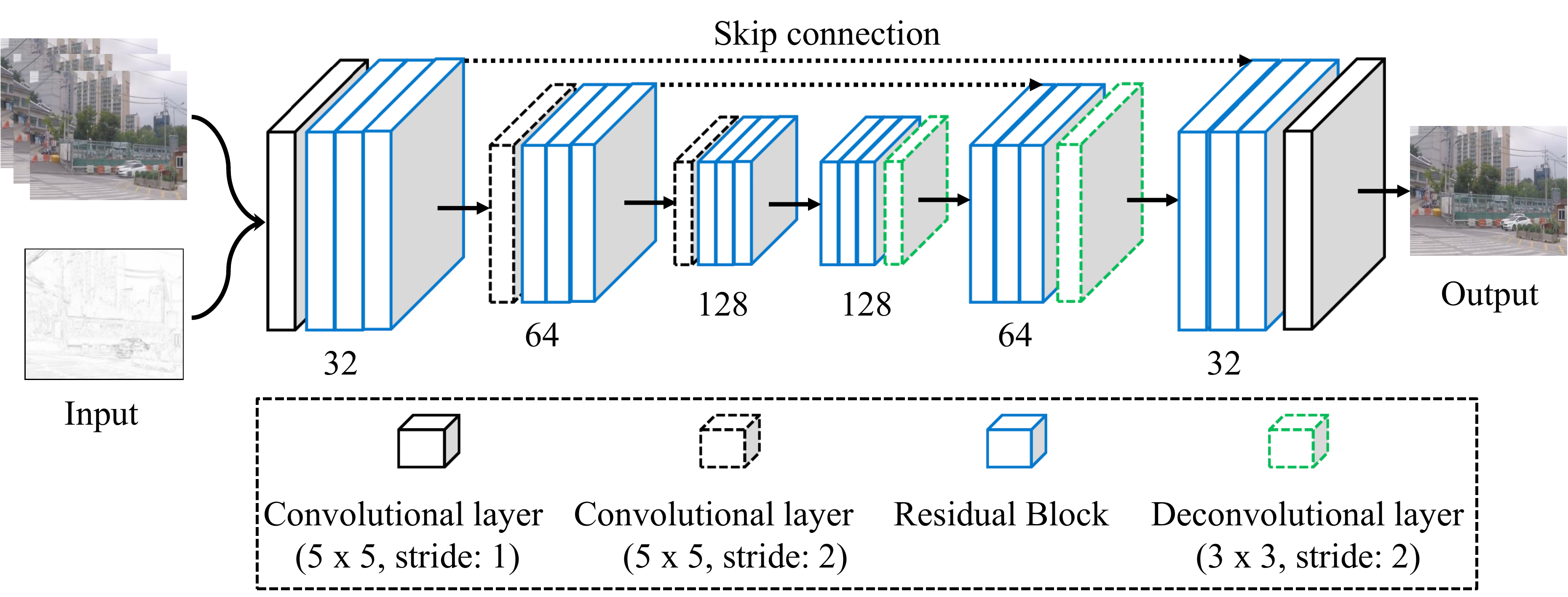}\\
\end{tabular}
\caption{Detailed network architectures for the latent frame restoration. Each convolutional and deconvolutional layers are followed by a ReLU unit except the last one that outputs the latent frame.
The number of feature channels in intermediate layers are $32$, $64$, $128$, $128$, $64$, and $32$, respectively.
}
\label{fig: restoration-flow-chart}
\vspace{-2mm}
\end{figure*}

%
We show the network details in this section. More detailed analysis and experimental results are included in the supplemental material which can be obtained at \url{https://jspan.github.io/}.
\vspace{-1mm}
{\flushleft \bf {Network details.}}
Figure~\ref{fig: flow-chart} shows the flowchart of the proposed algorithm at one stage. The proposed network shares the same network parameters when handling every three adjacent frames.
The network architecture for the latent image restoration is shown in Figure~\ref{fig: restoration-flow-chart}.
For the optical flow estimation, we use the PWC-Net~\cite{pwcnent/deqing} to estimate optical flow.
All the network modules are jointly trained in an end-to-end manner.


{\small
\bibliographystyle{ieee_fullname}
\bibliography{egbib}

\begin{thebibliography}{10}\itemsep=-1pt

\bibitem{burst/deblurring}
Miika Aittala and Fr{\'{e}}do Durand.
\newblock Burst image deblurring using permutation invariant convolutional
  neural networks.
\newblock In {\em ECCV}, pages 748--764, 2018.

\bibitem{variational/model/iccv07}
Leah Bar, Benjamin Berkels, Martin Rumpf, and Guillermo Sapiro.
\newblock A variational framework for simultaneous motion estimation and
  restoration of motion-blurred video.
\newblock In {\em ICCV}, pages 1--8, 2007.

\bibitem{Reblur2Deblur}
Huaijin Chen, Jinwei Gu, Orazio Gallo, Ming-Yu Liu, Ashok Veeraraghavan, and
  Jan Kautz.
\newblock Reblur2deblur: Deblurring videos via self-supervised learning.
\newblock In {\em ICCP}, pages 1--9, 2018.

\bibitem{cho/tog12/vd}
Sunghyun Cho, Jue Wang, and Seungyong Lee.
\newblock Video deblurring for hand-held cameras using patch-based synthesis.
\newblock {\em {ACM} TOG}, 31(4):64:1--64:9, 2012.

\bibitem{shengyangdai/cvpr08}
Shengyang Dai and Ying Wu.
\newblock Motion from blur.
\newblock In {\em CVPR}, 2008.

\bibitem{details/devil}
Jochen Gast and Stefan Roth.
\newblock Deep video deblurring: The devil is in the details.
\newblock In {\em ICCV Workshop}, 2019.

\bibitem{Gong/flow/blur}
Dong Gong, Jie Yang, Lingqiao Liu, Yanning Zhang, Ian~D. Reid, Chunhua Shen,
  Anton van~den Hengel, and Qinfeng Shi.
\newblock From motion blur to motion flow: {A} deep learning solution for
  removing heterogeneous motion blur.
\newblock In {\em CVPR}, pages 3806--3815, 2017.

\bibitem{kaiming/initialization}
Kaiming He, Xiangyu Zhang, Shaoqing Ren, and Jian Sun.
\newblock Delving deep into rectifiers: Surpassing human-level performance on
  imagenet classification.
\newblock In {\em ICCV}, pages 1026--1034, 2015.

\bibitem{FlowNet2/cvpr17}
Eddy Ilg, Nikolaus Mayer, Tonmoy Saikia, Margret Keuper, Alexey Dosovitskiy,
  and Thomas Brox.
\newblock Flownet 2.0: Evolution of optical flow estimation with deep networks.
\newblock In {\em CVPR}, pages 1647--1655, 2017.

\bibitem{STN}
Max Jaderberg, Karen Simonyan, Andrew Zisserman, and Koray Kavukcuoglu.
\newblock Spatial transformer networks.
\newblock In {\em NeurIPS}, pages 2017--2025, 2015.

\bibitem{kim/cvpr14/dynamic}
Tae~Hyun Kim and Kyoung~Mu Lee.
\newblock Segmentation-free dynamic scene deblurring.
\newblock In {\em CVPR}, pages 2766--2773, 2014.

\bibitem{VD/kim/cvpr15}
Tae~Hyun Kim and Kyoung~Mu Lee.
\newblock Generalized video deblurring for dynamic scenes.
\newblock In {\em CVPR}, pages 5426--5434, 2015.

\bibitem{DTBN/kim/iccv17}
Tae~Hyun Kim, Kyoung~Mu Lee, Bernhard Sch{\"{o}}lkopf, and Michael Hirsch.
\newblock Online video deblurring via dynamic temporal blending network.
\newblock In {\em ICCV}, pages 4058--4067, 2017.

\bibitem{sstn/eccv18}
Tae~Hyun Kim, Mehdi S.~M. Sajjadi, Michael Hirsch, and Bernhard
  Sch{\"{o}}lkopf.
\newblock Spatio-temporal transformer network for video restoration.
\newblock In {\em ECCV}, pages 111--127, 2018.

\bibitem{adam}
Diederik~P. Kingma and Jimmy Ba.
\newblock Adam: {A} method for stochastic optimization.
\newblock In {\em ICLR}, 2015.

\bibitem{li/cvpr10/vd}
Yunpeng Li, Sing~Bing Kang, Neel Joshi, Steven~M. Seitz, and Daniel~P.
  Huttenlocher.
\newblock Generating sharp panoramas from motion-blurred videos.
\newblock In {\em CVPR}, pages 2424--2431, 2010.

\bibitem{encoders/restoration}
Xiao-Jiao Mao, Chunhua Shen, and Yu-Bin Yang.
\newblock Image restoration using very deep convolutional encoder-decoder
  networks with symmetric skip connections.
\newblock In {\em NeurIPS}, pages 2802--2810, 2016.

\bibitem{lucky/frames/06}
Yasuyuki Matsushita, Eyal Ofek, Weina Ge, Xiaoou Tang, and Heung-Yeung Shum.
\newblock Full-frame video stabilization with motion inpainting.
\newblock {\em {IEEE} TPAMI}, 28(7):1150--1163, 2006.

\bibitem{KPN/cvpr18}
Ben Mildenhall, Jonathan~T. Barron, Jiawen Chen, Dillon Sharlet, Ren Ng, and
  Robert Carroll.
\newblock Burst denoising with kernel prediction networks.
\newblock In {\em CVPR}, pages 2502--2510, 2018.

\bibitem{MSCNN/deblur/cvpr17}
Seungjun Nah, Tae~Hyun Kim, and Kyoung~Mu Lee.
\newblock Deep multi-scale convolutional neural network for dynamic scene
  deblurring.
\newblock In {\em CVPR}, pages 257--265, 2017.

\bibitem{Nah/cvpr19}
Seungjun Nah, Sanghyun Son, and Kyoung~Mu Lee.
\newblock Recurrent neural networks with intra-frame iterations for video
  deblurring.
\newblock In {\em CVPR}, pages 8102--8111, 2019.

\bibitem{darkchannel/tpami18}
Jinshan Pan, Deqing Sun, Hanspeter Pfister, and Ming-Hsuan Yang.
\newblock Deblurring images via dark channel prior.
\newblock {\em {IEEE} TPAMI}, 40(10):2315--2328, 2018.

\bibitem{Encoders/inpainting}
Deepak Pathak, Philipp Kr{\"{a}}henb{\"{u}}hl, Jeff Donahue, Trevor Darrell,
  and Alexei~A. Efros.
\newblock Context encoders: Feature learning by inpainting.
\newblock In {\em CVPR}, pages 2536--2544, 2016.

\bibitem{DVD/cvpr17}
Shuochen Su, Mauricio Delbracio, Jue Wang, Guillermo Sapiro, Wolfgang Heidrich,
  and Oliver Wang.
\newblock Deep video deblurring for hand-held cameras.
\newblock In {\em CVPR}, pages 237--246, 2017.

\bibitem{pwcnent/deqing}
Deqing Sun, Xiaodong Yang, Ming-Yu Liu, and Jan Kautz.
\newblock {PWC-Net}: {CNNs} for optical flow using pyramid, warping, and cost
  volume.
\newblock In {\em CVPR}, pages 8934--8943, 2018.

\bibitem{SRN}
Xin Tao, Hongyun Gao, Xiaoyong Shen, Jue Wang, and Jiaya Jia.
\newblock Scale-recurrent network for deep image deblurring.
\newblock In {\em CVPR}, pages 8174--8182, 2018.

\bibitem{edvr}
Xintao Wang, Kelvin~C.K. Chan, Ke Yu, Chao Dong, and Chen Change~Loy.
\newblock {EDVR}: Video restoration with enhanced deformable convolutional
  networks.
\newblock In {\em CVPR Workshops}, 2019.

\bibitem{Wieschollek/vd/iccv17}
Patrick Wieschollek, Michael Hirsch, Bernhard Sch{\"{o}}lkopf, and Hendrik
  P.~A. Lensch.
\newblock Learning blind motion deblurring.
\newblock In {\em ICCV}, pages 231--240, 2017.

\bibitem{Wulff/eccv14/video/deblur}
Jonas Wulff and Michael~Julian Black.
\newblock Modeling blurred video with layers.
\newblock In {\em ECCV}, pages 236--252, 2014.

\bibitem{hardminingloss/inpainting}
Rui Xu, Xiaoxiao Li, Bolei Zhou, and Chen~Change Loy.
\newblock Deep flow-guided video inpainting.
\newblock In {\em CVPR}, pages 3723--3732, 2019.

\bibitem{zhang/3d/convolution}
Kaihao Zhang, Wenhan Luo, Yiran Zhong, Lin Ma, Wei Liu, and Hongdong Li.
\newblock Adversarial spatio-temporal learning for video deblurring.
\newblock {\em {IEEE} TIP}, 28(1):291--301, 2019.

\bibitem{zhoushanchen/iccv19}
Shangchen Zhou, Jiawei Zhang, Jinshan Pan, Haozhe Xie, Wangmeng Zuo, and Jimmy
  Ren.
\newblock Spatio-temporal filter adaptive network for video deblurring.
\newblock In {\em ICCV}, 2019.

\bibitem{epll}
Daniel Zoran and Yair Weiss.
\newblock From learning models of natural image patches to whole image
  restoration.
\newblock In {\em ICCV}, pages 479--486, 2011.

\end{thebibliography}
}

\end{document}